\documentclass[times, review, 10pt]{elsarticle}

\usepackage{amsmath}
\usepackage{amsthm}
\usepackage{amsfonts}
\usepackage{amssymb}
\usepackage{graphicx}
\usepackage[ruled,vlined]{algorithm2e}
\usepackage{enumitem}
\usepackage{bbm}
\usepackage{tabularx}
\usepackage{lineno,hyperref}
\usepackage{bm}
\usepackage{xcolor}
\usepackage{threeparttable}
\usepackage[caption=false,font=footnotesize]{subfig}
\usepackage{framed,multirow}
\usepackage{ragged2e}
\usepackage{makecell}
\usepackage{wrapfig}
\usepackage[top=4.3cm, right=4.8cm, bottom=4.3cm, left=4.8cm]{geometry}

\journal{Pattern Recognition}

\newcommand{\R}{\mathbb{R}}

\newsavebox\CBox
\def\textBF#1{\sbox\CBox{#1}\resizebox{\wd\CBox}{\ht\CBox}{\textbf{#1}}}

\makeatletter
\newcommand{\nosemic}{\renewcommand{\@endalgocfline}{\relax}}
\newcommand{\dosemic}{\renewcommand{\@endalgocfline}{\algocf@endline}}
\let\oldnl\nl
\newcommand{\nonl}{\renewcommand{\nl}{\let\nl\oldnl}}
\makeatother

\begin{document}
\begin{frontmatter}

\title{Optimal Multi-Kernel Combination for Online Multi-Label Classification}
\author[1]{Tingting~Zhai} \corref{cor1}   \ead{zhtt@yzu.edu.cn}
\author[1]{Wei~Liu}
\author[1]{Renjian~Yu}

\cortext[cor1]{Corresponding author}
\address[1]{College of Information and Artificial Intelligence, Yangzhou University, Yangzhou, China, 225127}

\begin{abstract}
Multi-label classification is particularly challenging in streaming scenarios, where data arrives sequentially.  
Online kernel methods offer a promising solution. 
However, prevailing methods rely on a single preselected kernel, making them prone to suboptimal choices since the global data distribution is unknown in advance in online settings. 
Although the recent method, Online Kernel Selection for Online Multi-label Classification (OKS-OMC), enables data-driven kernel selection, it fails to fully harness the complementary information from multiple kernels and encounters the curse of kernelization. To address these limitations, this paper introduces a novel online multi-kernel combination method for large-scale multi-label classification.
First, an efficient and robust online single-kernel multi-label method is designed.
Second, learning an optimal convex combination of multiple multi-label kernel classifiers is formalized as a structural risk minimization problem. An improved alternating optimization procedure is developed, in which the combination coefficients are updated by directly minimizing the objective function, while each kernel classifier is updated by minimizing a surrogate upper bound. This procedure decouples the learning of kernel classifiers from combination coefficients, allowing independent classifier updates via our single-kernel method and accelerating convergence to high-quality solutions. 
Our multi-kernel method is proven to asymptotically converge to the optimal fixed convex combination of the best fixed single-kernel models chosen in hindsight.
Ablation studies confirm that our method's components outperform their counterparts in OKS-OMC.
Comprehensive evaluations across 11 datasets and seven metrics demonstrate our method's overall superiority over existing online multi-label methods.
The source code of our method is available at \url{https://github.com/LUCKY-ting/OMKC}.
\end{abstract}


\begin{keyword}
	 online kernel methods  \sep  large-scale multi-label classification 
	 \sep data-driven kernel selection  \sep  multi-kernel learning \sep online learning
\end{keyword}

\end{frontmatter}


\section{Introduction}

Multi-label classification is a crucial and active subfield in machine learning, with broad applications spanning text categorization \cite{LiYPF24, YanLLXZ25}, image annotation \cite{CHEN2022, ZhangLLHZ26}, recommendation systems \cite{JainPY22, WuGLY25}, and bioinformatics \cite{DuH23, ArmahSekumSR24}.
Unlike traditional single-label problems, where each instance is assigned to exactly one category, multi-label classification requires models to simultaneously assign multiple relevant labels to each instance \cite{LiuWST22}.

Real-world applications now routinely generate massive multi-label data streams. On platforms such as Twitter and Sina Weibo, a continuous flow of posts—each associated with multiple labels —appears every second. Such streaming environments bring new challenges for classification.
First, it is infeasible to retrain offline models from scratch whenever new data arrives. 
Second, adapting offline methods to streaming data is far from straightforward. 

Online learning \cite{HoiSLZ21, ZhaiW23} presents a promising solution for streaming data. This paradigm continuously updates a model with incoming data, eliminating the need for full retraining. By integrating \emph{kernel methods} \cite{ScholkopfS02}, online methods can capture complex non-linear patterns, making them well-suited for the demands of multi-label stream classification.

Existing \emph{online kernel} methods for multi-label classification fall into two categories: \emph{single-kernel} and \emph{multi-kernel} methods.
Single-kernel methods rely on a preselected kernel function to implicitly map the input data into a high-dimensional feature space, where examples become approximately linearly separable and a linear classification algorithm can be applied. 
Within this category, the Passive Aggressive Multi-Label (PAML) family of algorithms \cite{ZhaiW23} has been introduced, extending the margin-based learning framework of SVMs to multi-label classification with the objective of maximizing classification margins. 
Further, the First-order Adaptive Label Thresholding (FALT) and First-order Fixed Label Thresholding (FFLT) methods \cite{ZhaiT2023} are proposed. These algorithms introduce distinct label thresholding strategies and reformulate the large-margin objective as an unconstrained problem, enabling joint optimization of the threshold predictor and all label scoring predictors via Online Gradient Descent (OGD).
In \cite{ZhaiH24}, FALT is further enhanced by optimizing the distribution of multi-label classification margins.
The main \emph{limitation} of single-kernel methods lie in that they rely on task-specific prior knowledge or an intensive cross-validation process to preselect a kernel function. However, in online learning scenarios where the global data structure is unknown in advance, this practice is vulnerable to \emph{suboptimal kernel selection}.

In contrast, \emph{multi-kernel} methods adopt a \emph{data-driven} approach to select appropriate kernel functions or combine different kernel classifiers during the online learning process. 
Currently, most existing online multi-kernel methods are designed for single-label classification tasks.
These methods can be divided into the following three categories: 
\begin{itemize}
\item  \emph{Online kernel selection methods}, which dynamically choose a kernel from a predefined set during the learning process. The approach in \cite{YangMJYH12} uses an exponentially weighted average to assign weights to candidate kernel classifiers, samples a kernel based on these weights, and updates the corresponding classifier with OGD. 
The approach was later enhanced in \cite{LiL22}, where kernel selection is formulated as a multi-armed bandit problem—with each kernel acting as an arm—and an adaptive parameter-tuning scheme is introduced to optimize the regret bound.
In a different strategy, \cite{ZhangL18} introduced an incremental sketched kernel alignment metric to select a kernel, also employing OGD to update the chosen classifier.

\item \emph{Online kernel combination methods}, which also leverage a predefined kernel pool but differ by updating multiple kernel classifiers and their combination coefficients in each round. 
Hoi et al. \cite{HoiJZY13} proposed to fuse two online algorithms: the Perceptron algorithm that learn individual classifiers, and the Hedge algorithm that assigns their weights. 
To simultaneously learn classifiers and their combination coefficients, Shen et al. \cite{ShenCG19} formulated a non-convex joint optimization problem, solving it through a fusion of OGD and the Exponentiated Gradient algorithm \cite{Hazan16}.
Further, Hong \cite{Hong24} introduced a collaborative Hedge algorithm to improve the combination coefficient update mechanism. To handle large kernel pools, Ghari and Shen \cite{GhariS23} proposed a graph-assisted framework that dynamically selects kernel subsets based on classifier losses or kernel similarities.

\item \emph{Methods that dynamically tune the kernel parameter}, which simultaneously learn the classifier and its kernel parameter during the learning process.
Nguyen et al. \cite{NguyenLBP17} adapted the Random Fourier Features (RFF) technique \cite{Rahimi2007} using the reparameterization trick. By lifting the randomness of Fourier components into an independent space,
their method allows for efficient, gradient-based online learning of the kernel parameter.
Zhang and Ning \cite{ZhangN20} transformed the learning problem into two optimal control problems. By solving both problems and following an alternating optimization scheme, the classifier parameters and the kernel bandwidth can be efficiently updated. 
Subsequently, \cite{ZhangNJT21} framed the entire learning task as a unified state feedback optimal control problem and employed an infinite‑horizon model predictive control technique to derive online update rules for both the classifier and the kernel bandwidth.
\end{itemize}

The aforementioned multi-kernel methods fail to capture the complex relationships between instances and labels that arise in \emph{online multi-label classification}.
A recent approach, Online Kernel Selection for Online Multi-label Classification (OKS-OMC) \cite{ZhaiL25}, attempts to bridge this gap. This method extends the single-kernel multi-label FALT algorithm \cite{ZhaiT2023} to a multi-kernel setting by formulating a joint optimization problem for multi-label classifiers and their combination coefficients, which is then approximately addressed by minimizing a tractable surrogate that upper-bounds the original objective. OKS-OMC is proved to converge to the best fixed single-kernel multi-label classifier chosen in hindsight. However, OKS-OMC is hampered by two critical drawbacks. Firstly, its regret bound indicates that it functions primarily as an online kernel selector, failing to fully harness the complementary information from multiple kernels. Secondly, it encounters the ``curse of kernelization" \cite{WangCV12, LuHWZL16}, as the size of each constituent model grows linearly with the number of data points, thereby compromising the scalability essential for online learning.

In this paper, we propose a novel online multi-kernel combination method for large-scale multi-label classification. Our contributions are as follows:
\begin{enumerate}
\item Firstly, we design a new online single-kernel multi-label classification method by integrating advanced RFF  techniques \cite{Rahimi2007, PenningtonYK15} with the fixed label thresholding method FFLT \cite{ZhaiT2023}. 
This integration offers a twofold advantage: 1) RFF ensures that the model size remains constant, independent of the data size; 2) The FFLT strategy enables robust learning in the high-dimensional feature spaces induced by RFF, effectively mitigating the overfitting risk in adaptively learning the label threshold.

\item Secondly, we formulate the learning of the optimal convex combination of multi-label kernel classifiers as a structural risk minimization problem. To solve it, we develop an improved alternating optimization procedure that decouples the learning of the kernel classifiers from that of the combination coefficients. This decomposition allows each kernel classifier to be updated seamlessly using our proposed single-kernel method, and facilitates fast convergence to a high-quality solution. We term the resulting method the Optimal Multi-Kernel Combination (OMKC) algorithm.

\item A theoretical analysis establishes that OMKC asymptotically converges to the \emph{optimal fixed} convex combination of the \emph{optimal fixed} single-kernel multi-label models chosen in hindsight. 
This indicates that OMKC can leverage the complementary information of multiple kernel functions.

\item Empirical results demonstrate the superiority of OMKC's coefficient update rule and classifier update strategy over those in OKS-OMC. Furthermore, comprehensive evaluations on 11 datasets using seven standard metrics demonstrate that OMKC statistically significantly outperforms existing online multi-label classification methods in terms of overall online performance. 
\end{enumerate}

The paper is structured as follows. Section \ref{sec.setting} details the notation and problem setting. Section \ref{sec.rff} introduces the RFF technique, which serves as a foundation for our method. Our method is elaborated in Section \ref{sec.algorithm}, alongside a theoretical analysis of its regret bound. Experimental results and discussions are presented in Section \ref{sec.experiments}. 
Finally, Section \ref{sec.conclusion} summarizes the study, discusses the limitations of our method, and suggests  future research directions.

\section{Notation and problem setting}
\label{sec.setting}

Let $(\bm x_1, Y_1), \cdots, (\bm x_T, Y_T)$ be a multi-label data stream, where at any time step $t$, the instance $\bm x_t \in \R^d$ is associated with a set of relevant labels $Y_t \subseteq \mathcal{Y}$. Here, $\mathcal{Y} = \{1, 2, \cdots, L\}$ is the full set of possible labels, and $\bar Y_t = \mathcal{Y}  \setminus Y_t$ denotes the set of irrelevant labels for $\bm x_t$.
We use the following mathematical notation throughout this paper.
The cardinality of a set $S$ is $\vert S \vert$. For a positive integer $N$,  $[N]$ denotes the set $\{1, 2, \cdots, N\}$.
The indicator function $\mathbbm{1}[a]$ outputs 1 if condition $a$ is true, and 0 otherwise.
The absolute value of a scalar $a$ is $|a|$. The $i$-th component of a vector $\bm u$ is $u_i$. 
The $\ell_1$, $\ell_2$ and $\ell_{\infty}$ norms of a vector $\bm u$ are denoted by $\Vert \bm u \Vert_1$, $\Vert \bm u \Vert$, and $\Vert \bm u \Vert_{\infty}$, respectively. The transpose of a matrix $\bm A$ is $\bm A^\top$ and its Frobenius norm is $\Vert \bm A \Vert_F$.

The online multi-label classification problem can be framed as a sequential prediction game.
In each round $t$ of this game, the environment reveals an instance $\bm x_t$ and requires the learner to predict a set of relevant labels $\hat Y_t \subseteq \mathcal{Y}$ for $\bm x_t$. After the prediction, the true relevant label set $Y_t$ is disclosed and the learner incurs a loss that quantifies its prediction performance. Based on this loss, the learner then refines its model, aiming to minimize the cumulative loss over all rounds, which is equivalent to maximizing the sequential multi-label prediction performance.

\section{Random Fourier Features (RFF)}
\label{sec.rff}

RFF \cite{Rahimi2007} is a technique to approximate shift-invariant kernels.
Its core idea is to find an explicit feature mapping $\hat{\phi}(\cdot)$ with a user-controllable dimension, such that the inner product $\hat{\phi}(\bm{x})^\top \hat{\phi}(\bm{z})$ approximates the kernel function $\mathcal{K}(\bm{x}, \bm{z})$.  
A kernel is shift-invariant if it satisfies $\mathcal{K}(\bm{x}, \bm{z}) = K(\bm{\delta})$ for $\bm{\delta} = \bm{x} - \bm{z}$ and some function $K$.
Given a shift-invariant kernel ${K}(\cdot)$, its Fourier transform, denoted by $p(\bm u)$, can be expressed as
\begin{align}
p(\bm u) = (\frac{1}{2 \pi})^d \int_{\R^d} K(\bm\delta) \exp\bigl(-j {\bm\delta}^\top {\bm u}\bigr) d {\bm\delta}.
\label{ft}
\end{align}

Bochner's theorem guarantees that $p(\bm u)$ can be viewed as a probability distribution if ${K}(\bm\delta)$ is properly scaled \cite{Rahimi2007}. According to Fourier inversion theorem, $K(\bm\delta)$ can be recovered from its Fourier transform  $p(\bm u)$:
\begin{align*}
K(\bm\delta) = \int_{\R^d} p(\bm u) \exp \bigl( j {\bm u}^\top {\bm\delta}\bigr) d {\bm u} 
= \mathbb{E}_{\bm u \sim p(\bm u)} [\exp \bigl(j {\bm u}^\top {\bm\delta}\bigr)]  
= \mathbb{E}_{\bm u \sim p(\bm u)} [\cos({\bm u}^\top {\bm\delta}) ]
\end{align*}
where the last equation results from neglecting the imaginary part, $j \sin({\bm u}^\top {\bm\delta})$.
Thus, we obtain 
$\mathcal{K}(\bm x, \bm z) = \mathbb{E}_{\bm u \sim p(\bm u)} [ \cos({\bm u}^\top {(\bm x - \bm z}))] $.
By using the trigonometric identity
\begin{align*}
 \cos({\bm u}^\top {(\bm x - \bm z})) 
= \cos({\bm u}^\top \bm x +  b) \cos({\bm u}^\top \bm z +  b) +\sin({\bm u}^\top \bm x +  b) \sin({\bm u}^\top \bm z +  b)
\end{align*}
and defining $\varphi_{\bm u, b}(\bm x) = [\cos({\bm u}^\top \bm x +  b) , \sin({\bm u}^\top \bm x +  b)] ^\top$,  we get that $ \cos({\bm u}^\top {(\bm x - \bm z})) = \varphi_{\bm u, b}(\bm x) ^\top  \varphi_{\bm u, b}(\bm z) $.
Let $\mathcal{U}(0, 2\pi)$ denote a uniform distribution in $[0, 2\pi]$.
Taking the expectation over $\bm u \sim p(\bm u)$ and $b \sim \mathcal{U}(0, 2\pi)$ yields
\begin{align}
\mathcal{K}(\bm x, \bm z) 
&=\mathbb{E}_{\bm u \sim p(\bm u),\ b \sim \mathcal{U}(0, 2\pi)} [\varphi_{\bm u, b}(\bm x) ^\top  \varphi_{\bm u, b}(\bm z)] .
\label{mid2}
\end{align}

By randomly drawing $D$ independent samples $\bm u_1, \bm u_2, \cdots, \bm u_D$ from $p(\bm u)$, and also $b_1, b_2, \cdots, b_D$ from $\mathcal{U}(0, 2\pi)$, the expectation in Eq.(\ref{mid2}) can be approximated by the sample average:
\begin{align}
\mathcal{K}(\bm x, \bm z) 
\approx  \frac{1}{D} \sum_{k=1}^D  \varphi_{\bm u_k, b_k}(\bm x) ^\top  \varphi_{\bm u_k, b_k}(\bm z) 
     = \hat \phi (\bm x) ^\top \hat \phi (\bm z) 
\label{mid3}
\end{align}
where $\hat \phi (\bm x) $ is an explicit feature map of $2D$ dimensionality, defined as
\begin{align} 
\hat\phi (\bm x) = \frac{1}{\sqrt{D}} [\varphi_{\bm u_1, b_1}(\bm x)^\top,  \cdots,  \varphi_{\bm u_D, b_D}(\bm x)^\top ]^\top .
\label{fmap}
\end{align}

According to Eq.(\ref{mid3}), $\hat \phi (\bm x) ^\top \hat \phi (\bm z) $ is unbiased estimate of $\mathcal{K}(\bm x, \bm z) $. Further, it is proven that their difference decreases at an exponential rate in $D$ \cite{Rahimi2007}.

To construct the feature map $\hat \phi (\cdot) $, we must calculate out $p(\bm u)$ by Eq.(\ref{ft}) and then
sample $\bm u$ from $p(\bm u)$.
A computational simplification arises when the kernel function is separable.
If a kernel function satisfies $K(\bm\delta) = \prod_{i=1}^d k(\delta_i)$, then its Fourier transform $p(\bm u)$ factorizes accordingly: $p(\bm u) = \prod_{i=1}^d p(u_i)$. 
This factorization implies that all components of $\bm u$, that is, $u_1, u_2,\cdots, u_d$, are independent and identically distributed, each following a one-dimensional marginal distribution $p(u_i)$.
Thus, to draw a $d$-dimensional vector $\bm u$ from $p(\bm u)$, one simply draws $d$ independent samples from the univariate distribution $p(u_i)$.
Table \ref{rff} lists several common shift-invariant kernels, their Fourier transforms, and the corresponding sampling distribution for $u_i$. 

\begin{table}[!htp]
	\renewcommand{\arraystretch}{1}
	\caption{Shift-invariant kernels, their Fourier transforms and the corresponding distribution for $u_i$.} 
	\label{rff}
	\footnotesize
	\centering \tabcolsep =2pt
	\begin{tabular}{cccc}
	\hline
	Kernel name &$K(\bm\delta)$  &$p(\bm u)$   &distribution of $u_i$ \\
	\hline  
       Gaussian      &$\exp \bigl(-\frac{||\bm\delta||^2}{2 \sigma^2} \bigr)$     &$\prod_{i=1}^d \frac{\sigma}{\sqrt{2 \pi}} \exp \bigl(-\frac{\sigma^2 u_i^2}{2}\bigr)$          & Gaussian $\mathcal{N}(0, 1/\sigma^2)$    \\
       Laplacian     &$\exp\bigl(-\frac{||\bm\delta||_1}{\sigma}\bigr)$     &$\prod_{i=1}^d \frac{\sigma}{\pi (1 + \sigma^2 u_i^2)}$          
       & Cauchy $\mathcal{C}(0, 1/\sigma)$    \\
       Cauchy  &$\frac{1}{2^d}\prod_{i=1}^d \frac{2}{1+(\delta_i/\sigma)^2}$  &$\prod_{i=1}^d \frac{\sigma}{2} \exp \bigl(-\sigma |u_i| \bigr)$  
       &Laplace  $\mathcal{L}(0, 1/\sigma)$\\
       	\hline  
	\end{tabular}
\end{table}
 
The introduction of RFF has spurred extensive research into kernel approximation techniques; we refer to \cite{LiuHCS22} for a comprehensive survey. While initially developed for shift-invariant kernels, the RFF framework has since been successfully generalized to other kernel classes, including arc-cosine kernels \cite{LiuHCS22, ChoS09} and polynomial kernels \cite{PenningtonYK15, MeisterSW19}.

\section{The proposed approach}
\label{sec.algorithm}

\subsection{RFF-based online single-kernel multi-label classification}

Given a Mercer kernel $\mathcal{K}_p$,  Mercer's theorem \cite{ScholkopfS02} guarantees the existence of an implicit feature map $\phi_p(\cdot)$ to a reproducing kernel Hilbert space $\mathcal{H}_p$ such that this kernel can be expressed as an inner product in this space: $\mathcal{K}_p(\bm x, \bm z) = \langle \phi_p(\bm x), \phi_p(\bm z) \rangle_{\mathcal{H}_p}$. We can use this map $\phi_p(\cdot)$ to transform instances from the original input space to $\mathcal{H}_p$, thereby learning our multi-label classifier in $\mathcal{H}_p$.

Our single-kernel multi-label classifier is constructed based on the First-order Fixed Label Thresholding (FFLT) method \cite{ZhaiT2023}.
Specifically, the classifier maintains $L$ predictors, each generating a score for one of the $L$ labels, coupled with a predefined global fixed threshold $\gamma$ to discriminate between relevant and irrelevant labels. Formally, the predictors are parameterized by a matrix $\bm W_p = [\bm w_p^{(1)}, \cdots, \bm w_p^{(L)}]$, where for any $i \in [L]$, $\bm w_p^{(i)}\in \mathcal{H}_p$ corresponds to the scoring predictor for $i$-th label.
For an instance $\bm x_t$, its new feature representation in $\mathcal{H}_p$ is $\phi_p(\bm x_t)$.
The set of relevant labels for $\bm x_t$ is then predicted as 
$\{ i \in \mathcal{Y}: \phi_p(\bm x_t)^\top \bm w_p^{(i)} > \gamma \}$.
The rule classifies a label as relevant if its score $\phi_p(\bm x_t)^\top \bm w_p^{(i)}$ exceeds the threshold $\gamma$. The rule directly implies the following loss function for evaluating performance:
\begin{align*}
\ell (\bm W_p; (\phi_p(\bm x_t), Y_t)) 
& =  \frac{1}{|Y_t|} \sum_{i \in Y_t} \max \left\{0,   1 -  \left(\phi_p(\bm x_t)^\top \bm w_p^{(i)} - \gamma \right)\right\} \nonumber\\
& +  \frac{1}{|\bar Y_t|} \sum_{i \in \bar Y_t} \max \left\{0,  1 - \left(\gamma -  \phi_p(\bm x_t)^\top \bm w_p^{(i)} \right)\right\}.
\end{align*}
Note that $\ell (\bm W_p; (\phi_p(\bm x_t), Y_t))$ consists of a sum of losses over all labels.
The loss associated with each label is weighted according to the contribution of the current instance to that label. 
For a relevant label, the contribution degree is quantified as $\frac{1}{|Y_t|}$, whereas for an irrelevant label, it is quantified as $\frac{1}{|\bar{Y}_t|}$.
This specific weighting mechanism ensures that within each instance, relevant and irrelevant labels contribute equally to the loss, preventing either side from dominating the learning signal.

Our single-kernel multi-label model is learnt by solving the following structural risk minimization problem:
\begin{align}
\min_{\bm W_p}
\left\{ \frac{1}{T} \sum_{t=1}^T f (\bm W_p; (\phi_p(\bm x_t), Y_t)) \right\}
\label{orig-single}
\end{align}
where 
$ f(\bm W_p; (\phi_p(\bm x_t), Y_t)) =\ell (\bm W_p; (\phi_p(\bm x_t), Y_t))  + \frac{\lambda}{2} \Vert \bm W_p \Vert_F^2 $, and the hyperparameter $\lambda > 0$ is used to control the model complexity.

Problem (\ref{orig-single}) is convex. Thus in an online setting, it can be solved using OGD. 
For convenience, let $f_t(\bm W_p) $ be the shorthand for $f(\bm W_p; (\phi_p(\bm x_t), Y_t))$.
Let $\bm W_{p,t} = [\bm w_{p,t}^{(1)}, \cdots,  \bm w_{p,t}^{(L)}] $ be the single-kernel multi-label model at online round $t$.
Following the steps of OGD,  $\bm w_{p, t}^{(i)}$ is first initialized to zero vector for any $i$, then it can be incrementally updated as 
\begin{align}
\bm w_{p, t+1}^{(i)} = \bm w_{p, t}^{(i)} -  \eta \nabla_{\bm w_{p}^{(i)}} f_t(\bm W_{p, t}), \ \forall i \in [L], 
\label{ogd}
\end{align}
where 
$\nabla_{\bm w_{p}^{(i)}} f_t(\bm W_{p, t})$ belongs to the sub-gradient set of $f_t(\bm W_{p})$ with respect to $\bm w_{p}^{(i)}$ at $\bm W_{p, t}$, and $\eta < 1/ \lambda$ is the learning rate.
Given that $\nabla_{\bm w_{p}^{(i)}} f_t(\bm W_{p, t}) = \lambda \bm w_{p, t}^{(i)} + \gamma_{p, t}^{(i)} \phi_p(\bm x_t)$ where 
\begin{equation}
\gamma_{p, t}^{(i)}  =
\begin{cases}
-\frac{\mathbbm{1} [\phi_p(\bm x_t)^\top  \bm w_{p,t}^{(i)}  - \gamma < 1]}{|Y_t|} 		    &\mbox { if } i \in Y_t, \\
\ \frac{\mathbbm{1} [\gamma - \phi_p(\bm x_t)^\top \bm w_{p,t}^{(i)} < 1]}{|\bar Y_t|}       &\mbox { if } i \in \bar Y_t, \\
\end{cases}
\label{detail}
\end{equation}
the update formula (\ref{ogd}) of OGD can be rewritten as
\begin{equation}
\label{update1}
\bm w_{p, t+1}^{(i)} = (1 - \lambda \eta) \bm w_{p, t}^{(i)} - \eta \gamma_{p, t}^{(i)} \phi_p(\bm x_t) , \ \forall i \in [L].
\end{equation}

A recursive expansion of Eq.({\ref{update1}}) yields 
 $\bm w_{p, t+1}^{(i)} = \sum_{k=1}^t  \alpha_{p,k}^{(i)} \phi_p (\bm x_k)$, where $\alpha_{p,k}^{(i)}$ is the coefficient corresponding to $\phi_p (\bm x_k)$.
For many popular kernels, such as Gaussian kernels, the map $\phi_p(\cdot)$ lacks an explicit, finite-dimensional form. 
Consequently, the representation of these kernel-based scoring predictors grows linearly with the number of observed examples, incurring prohibitive computational and storage costs on large-scale data streams. This well-known scalability problem is termed the "curse of kernelization" \cite{WangCV12}.

RFF-based kernel approximation techniques provide a breakthrough for this problem by constructing an explicit, approximate feature map $\hat \phi_p(\cdot)$ that satisfies $\mathcal{K}_p(\bm x, \bm z) \approx \hat \phi_p(\bm x)^\top \hat \phi_p(\bm z)$. 
Using $\hat\phi_p(\cdot)$, instances are mapped to a finite-dimensional feature space, denoted by $\hat{\mathcal{H}}_p$.
Substituting $\hat{\phi}_p(\cdot)$ from Eq.(\ref{fmap}) for the implicit map ${\phi}_p(\cdot)$ in Eq.(\ref{detail}) and (\ref{update1}), we obtain
\begin{equation}
\label{key-update1}
\bm w_{p, t+1}^{(i)} = (1 - \lambda \eta) \bm w_{p, t}^{(i)} - \eta \gamma_{p, t}^{(i)} \hat{\phi}_p(\bm x_t) , \ \forall i \in [L].
\end{equation}

Now, each predictor $\bm w_{p, t+1}^{(i)}$ can be represented and stored as a fixed-dimensional vector in $\R^{2D}$.
Consequently, the task of learning a single-kernel multi-label model is reduced to learning a linear multi-label model in $\R^{2D}$, thereby perfectly combining the efficiency of the linear learning algorithm with the expressive power of nonlinear kernels.

\subsection{Multi-kernel multi-label classification}

Single-kernel methods depend heavily on a manually selected kernel, a process that is both computationally expensive and prone to suboptimal choices, especially with limited data. Online multi-kernel learning overcomes this by automating kernel selection and combination during the learning process.

Let $\mathcal{K}_1, \cdots, \mathcal{K}_P$ denote $P$ predefined Mercer kernels, with $\hat{\phi}_1(\cdot), \cdots, \hat{\phi}_P(\cdot) $ being their respective RFF approximate feature maps. Thus, an instance $\bm{x}_t $ is projected into $P$ feature spaces, yielding representations $\hat{\phi}_1(\bm{x}_t), \cdots, \hat{\phi}_P(\bm{x}_t)$.

For each kernel $\mathcal{K}_p$, let $\bm{W}_p$ denote its associated single-kernel multi-label model, constructed as described in the previous subsection. Many methods exist for combining these single-kernel models. Here, we adopt the widely used convex combination strategy. Specifically, we assign a non-negative coefficient $\mu_p$ to each model $ \bm{W}_p$, subject to $\sum_{p=1}^P \mu_p = 1$.
Let $\mathcal{W} = \{\bm W_1, \cdots,\bm W_P\}$ and $\bm \mu = [\mu_1, \cdots, \mu_P]^\top \in \Delta$, where 
$\Delta$ denotes a probability simplex in $\R^P$, defined as $\Delta = \{ \bm r \in [0, 1]^P :  \sum_{p=1}^P r_p =1 \}$.
Our multi-kernel multi-label model is parameterized by $(\mathcal{W}, \bm \mu)$.

Under this convex combination scheme, the score for the $i$-th label of an instance $\bm{x}_t$ is computed as the convex combination of predictions from all single-kernel label scoring predictors:
$h^{(i)}(\bm{x}_t) = \sum_{p=1}^P \mu_p \hat{\phi}_p(\bm{x}_t)^\top \bm{w}_p^{(i)}$.
The set of relevant labels for $\bm{x}_t$ is then predicted as:
$\hat{Y}_t = \left\{ i \in \mathcal{Y} : h^{(i)}(\bm{x}_t) > \gamma \right\}$,  
where $\gamma$ is the predefined threshold. 
This decision rule naturally gives rise to the following loss function for evaluating the performance of our multi-kernel multi-label model:
\begin{align*}
\mathcal{L} (\mathcal{W}, \bm \mu; (\bm x_t, Y_t)) 
&= \frac{1}{|Y_t|} \sum_{i \in Y_t} \max \left\{0,   1 -  \left( \sum_{p=1}^P \mu_p \hat{\phi}_p(\bm x_t)^\top \bm w_p^{(i)}  - \gamma \right) \right\} \nonumber \\
&+ \frac{1}{|\bar Y_t|} \sum_{i \in \bar Y_t} \max \left\{0,  1 -  \left(\gamma - \sum_{p=1}^P \mu_p \hat{\phi}_p(\bm x_t)^\top \bm w_p^{(i)}\right)\right\}.
\end{align*}

In the framework of structural risk minimization, we hope to find $(\mathcal{W}, \bm \mu)$ such that the regularized loss is minimized:
\begin{align}
\min_{\mathcal{W}, \bm \mu}
\left\{\frac{1}{T} \sum_{t=1}^T \mathcal{F}  (\mathcal{W}, \bm \mu; (\bm x_t, Y_t)) \right\} 
\label{orig}
\end{align}
where 
\begin{align*}
\mathcal{F} (\mathcal{W}, \bm \mu; (\bm x_t, Y_t)) =
\mathcal{L} (\mathcal{W}, \bm \mu; (\bm x_t, Y_t)) + \frac{\lambda}{2} \sum_{p=1}^P \mu_p \Vert \bm W_{p} \Vert_F^2.
\end{align*}

The hyperparameter $\lambda > 0$ controls the strength of regularization. The combination coefficients are also embedded within the regularization term, which scales the penalty in a capacity-aware manner \cite{ZhaiL25}.

The optimization problem (\ref{orig}) is convex with respect to $\mathcal{W}$ alone and $\bm \mu$ alone, but it is \emph{non-convex} in the joint variable $(\mathcal{W}, \bm \mu)$. This lack of joint convexity presents a primary challenge. A secondary challenge arises from its scale: the problem has $2DLP+P-1$ parameters, which grows linearly with $P$. For large $P$, this significantly exceeds the $2DL$ parameters of the single-kernel problem (\ref{orig-single}).

Finding the global optimum of (\ref{orig}) is computationally expensive, even in an offline setting—let alone in an online scenario. Therefore, we focus on identifying a strong local optimum that guarantees robust performance on newly arriving data, rather than pursuing the globally optimal solution.

A natural approach to solve an optimization problem with two variable blocks is the Generalized Alternating Minimization (GAM) \cite{JainK17}, which iteratively fixes one variable block and optimizes the other by minimizing the original objective function. If closed-form solutions for the marginal optimizations are unavailable, gradient descent steps can be applied instead. However, in an online setting, GAM suffers from extremely slow convergence, as confirmed in Section \ref{evidence}.
To address this, we propose an improved alternating optimization strategy, updating the variables from $(\mathcal{W}_t, \bm{\mu}_t)$ to $(\mathcal{W}_{t+1}, \bm{\mu}_{t+1})$ as follows:
\begin{enumerate}[label=Step \arabic*:, leftmargin=*]
\item fix $\mathcal{W}$ to $\mathcal{W}_t$ and obtain $\bm{\mu}_{t+1}$ by minimizing the original objective function 
$\mathcal{F} (\mathcal{W}_t, \bm \mu; (\bm x_t, Y_t))$;
\item fix $\bm{\mu}$ to $\bm{\mu}_{t+1}$ and obtain $\mathcal{W}_{t+1}$ by minimizing \emph{an upper bound of the original objective} $\mathcal{F} (\mathcal{W}, \bm \mu_{t+1}; (\bm x_t, Y_t))$.
\end{enumerate}

Crucially, optimizing an upper bound in Step 2 effectively decouples the $\mathcal{W}$-update from $\bm{\mu}$, permitting independent optimization of $\mathcal{W}$, as demonstrated later.  
This strategy facilitates faster convergence to a high-quality solution.

When $\mathcal{W}$ is fixed to $\mathcal{W}_t$ in Step 1, the marginal minimization problem with respect to $\bm{\mu}$ lacks a closed-form solution. We therefore perform a gradient descent on $\bm{\mu}$.
Since $\bm{\mu}$ is constrained to the probability simplex, we apply the Exponential Gradient Descent (EGD) \cite{hazan2023} to maintain feasibility. 

Specifically, define $G_t(\bm{\mu}) = \mathcal{F} (\mathcal{W}_t, \bm \mu; (\bm x_t, Y_t))$ and let $\nabla_{\bm \mu} G_t(\bm \mu_t)$ denote its gradient at $\bm \mu_t$. The $p$-th component of this gradient, denoted by $[\nabla_{\bm \mu} G_t(\bm \mu_t)]_p$, is computed as:
\begin{align}
 [\nabla_{\bm \mu} G_t(\bm \mu_t)]_p =  \frac{\lambda}{2} \Vert \bm W_{p, t} \Vert_F^2
 - \frac{1}{|Y_t|}\sum_{i \in Y_t} a_t^{(i)} \hat{\phi}_p(\bm x_t)^\top \bm w_{p, t}^{(i)}  \nonumber\\
 + \frac{1}{|\bar Y_t|} \sum_{i \in \bar Y_t} b_t^{(i)} \hat{\phi}_p(\bm x_t)^\top \bm w_{p, t}^{(i)} 
 \label{grad}
\end{align}
where $a_t^{(i)} = \mathbbm{1}[\sum_{p=1}^P \mu_{p, t} \hat{\phi}_p(\bm x_t)^\top \bm w_{p, t}^{(i)}  - \gamma < 1]$
and $b_t^{(i)} = \mathbbm{1}[\gamma - \sum_{p=1}^P \mu_{p, t} \hat{\phi}_p(\bm x_t)^\top \bm w_{p, t}^{(i)} < 1] $.

Thus, each component of $\bm\mu_t = [\mu_{1, t}, \cdots, \mu_{P, t}]^\top$ can be incrementally updated as:
\begin{align}
\mu_{p, t+1} = \frac{\mu_{p, t} \exp{\left( -\beta \, [\nabla_{\bm \mu} G_t(\bm \mu_t)]_p \right)}}{\sum_{q=1}^P \mu_{q, t} \exp{\left( -\beta \, [\nabla_{\bm \mu} G_t(\bm \mu_t)]_q \right)}},  \forall p \in [P]
\label{key-update2}
\end{align}
where $\beta > 0$ is the learning rate for this update.

In Step 2, we consider the following upper bound of $\mathcal{F} (\mathcal{W}, \bm \mu_{t+1}; (\bm x_t, Y_t))$: 
\begin{align}
\mathcal{F} (\mathcal{W}, \bm \mu_{t+1}; (\bm x_t, Y_t)) 
\leq_1 &\sum_{p=1}^P \mu_{p, t+1} \ell (\bm W_p; (\hat\phi_p(\bm x_t), Y_t)) + \frac{\lambda}{2} \sum_{p=1}^P \mu_{p, t+1} \Vert \bm W_{p} \Vert_F^2   \nonumber\\
 = &   \sum_{p=1}^P \mu_{p, t+1} \left(  \ell (\bm W_p; (\hat\phi_p(\bm x_t), Y_t)) + \frac{\lambda}{2} \Vert \bm W_{p} \Vert_F^2 \right)  \nonumber\\
 = & \sum_{p=1}^P \mu_{p, t+1}   f(\bm W_p; (\hat\phi_p(\bm x_t), Y_t)) 
\label{upperbound}
\end{align}
where $\leq_1$ follows from applying Jensen's inequality $g(\sum_{p=1}^P \mu_p z_p) \leq \sum_{p=1}^P \mu_p g(z_p)$  \cite{BV2004} to the convex function $g(z) = \max \{0, 1- z\}$.

Eq. (\ref{upperbound}) suggests that $\mathcal{F} (\mathcal{W}, \bm \mu_{t+1}; (\bm x_t, Y_t))$ is upper bounded by a convex combination of single-kernel losses $ f(\bm W_p; (\hat\phi_p(\bm x_t), Y_t))$.
Minimizing this upper bound over $\mathcal{W}$ thus reduces to independently minimizing each $ f(\bm W_p; (\hat\phi_p(\bm x_t), Y_t))$ over $\bm W_p$. As this marginal optimization problem also lacks a closed-form solution,  we apply a gradient descent on each $\bm W_p$.
Consequently, each $\bm W_{p, t+1}$ in $\mathcal{W}_{t+1}=\{\bm W_{1,t+1}, \cdots, \bm W_{P,t+1}\}$ can be obtained directly via Eq. (\ref{key-update1}). 
Notably, this optimization over $\mathcal{W}$ is independent of $\bm{\mu}$.

To summarize, each round $t$ involves updating the combination coefficients with Eq. (\ref{key-update2}) and 
updating each single-kernel multi-label model with Eq. (\ref{key-update1}).

We call this proposed approach the \emph{Optimal Multi-Kernel Combination algorithm (OMKC)}. This name reflects its capability to optimally combine multiple single-kernel multi-label classifiers, a property we formally demonstrate in the next section. The pseudocode for OMKC is provided in Algorithm \ref{ours}.

\begin{algorithm} 
	\LinesNumbered
	\caption{OMKC}
	\label{ours}
	\nosemic
	\KwIn{$\lambda > 0,  0 < \eta < 1 /\lambda,  \beta > 0, \gamma, D, \{\mathcal{K}_1, \mathcal{K}_2, \cdots, \mathcal{K}_P\}$}
	\KwOut{$\bm \mu_{T+1}$, $\bm{W}_{p,T+1}$ for any $p \in [P]$}
	$\bm {\mu}_1 = [{\mu}_{1,1}, \cdots, {\mu}_{P,1}]^\top = [1/P, \cdots, 1/P]^\top$;
	$\bm{W}_{p,1} = [\bm w_{p,1}^{(1)}, \cdots, \bm w_{p,1}^{(L)}] = [\mathbf{0}, \cdots, \mathbf{0}]$, $\forall p \in [P]$\; 
	\For {$p=1, 2, \cdots, P$}
	{
	Obtain $p(\bm u)$ for the kernel $\mathcal{K}_p$ using Eq.(\ref{ft}) \;
	Draw $\bm u_1, \bm u_2, \cdots, \bm u_D$ from $p(\bm u)$, and $b_1, b_2, \cdots, b_D$ from $\mathcal{U}(0, 2\pi)$\;
	Obtain the explicit feature map ${\hat\phi}_p (\cdot)$ associated with the kernel $\mathcal{K}_p$ via Eq.(\ref{fmap})\;
	}
	\For {$t=1, 2, \cdots, T$}
	{
	Receive $\bm{x}_t \in \R^d$\; 
	Calculate ${\hat\phi}_1 (\bm{x}_t), {\hat\phi}_2 (\bm{x}_t), \cdots,  {\hat\phi}_P (\bm{x}_t)$\;
	Predict: 
	$\hat{Y}_t = \{ i \in \mathcal{Y}: 
	\sum_{p=1}^P \mu_{p, t} {\hat\phi}_p(\bm x_t)^\top \bm w_{p, t}^{(i)} 
	 > \gamma  \}$\;
	Receive $Y_t \subseteq \mathcal{Y}$\;
	\For {$p=1, 2, \cdots, P$}
		{
		Update ${\mu}_{p, t+1}$ by Eq. (\ref{key-update2}) \;
		Update $\bm W_{p, t+1}$ by Eq. (\ref{key-update1}) \;
		}
	}
\end{algorithm}

According to Algorithm \ref{ours}, we can analyze the space and time complexity of OMKC.
The total space complexity of OMKC comprises two components: the storage of RFF components and the storage of multi-label models. For each of the $P$ kernels, OMKC maintains $D$ RFF components, and storing each component requires $O(d)$ space, resulting in a complexity of $O(DPd)$. Additionally, OMKC maintains $P$ multi-label models, each containing $L$ label scoring predictors. Since each predictor requires $O(2D)$ space, this part contributes a complexity of $O(2DPL)$. Therefore, the overall space complexity is $O(DP(d + 2L))$.

The total time complexity for OMKC arises from two parts: computing the RFF mappings (Steps 2-5) and performing online prediction and model updates (Step 6-13). 
The first part involves sampling $D$ RFF components $P$ times, with each component requiring $O(d)$ time, resulting in a complexity of $O(DPd)$. However, this computation is data-independent and can be performed in advance as a pre-processing step. Therefore, its time complexity is excluded from the total online complexity.
Regarding the second part, we first analyze its per-round complexity.
In each round, computing RFF feature vectors (Step 8) costs $O(dDP)$, online prediction (Step 9) costs $O(2DPL)$, and updating the multi-kernel multi-label model (Steps 11-13) costs $O(2DPL + PL) = O(2DPL)$.
Thus, the per-round time complexity is $O(DP(d+4L))$, leading to a total complexity $O(DP(d+4L)T)$.

Crucially, the space and per-round time complexities of OMKC do not grow with the length of the data stream. This allows OMKC to circumvent the curse of kernelization.

\subsection{Theoretical analysis}

Since OMKC uses RFF feature maps satisfying $\Vert\hat\phi_p(\bm x_t) \Vert= 1$ for all $t \in [T]$ and $p \in [P]$, we can extend the analysis of Lemma 1 in \cite{ZhaiL25} to OMKC.
This yields the boundedness for each label scoring predictor $\bm w_{p, t}^{(i)}$ and each single-kernel multi-label model $\bm W_{p,t}$:  
 $||\bm w_{p, t}^{(i)}|| \leq 1 / \lambda$ and $\Vert \bm W_{p,t} \Vert_F \leq \sqrt{2}/ \lambda$ for all $p, t, i$.
Based on these facts, we can prove that the classifier updating strategy of OMKC satisfies, for any $p \in [P]$,
\begin{equation}  \label{regret2}
\sum_{t=1}^T f(\bm W_{p,t}; (\hat\phi_p(\bm x_t), Y_t)) -  \min_{\bm W_{p}} \sum_{t=1}^T f(\bm W_{p}; (\hat\phi_p(\bm x_t), Y_t)) 
\leq O(\sqrt{T}).
\end{equation}

This sublinear regret bound (\ref{regret2}) implies that each single-kernel multi-label model maintained by OMKC asymptotically converges to the \emph{optimal fixed} model chosen in hindsight from the corresponding kernel-induced RFF feature space.

Next, we prove that the gradient $\nabla_{\bm \mu} G_t(\bm \mu_t)$ used in the EGD step is bounded, which is essential for deriving the theoretical guarantee for OMKC's coefficient update strategy.
According to Eq.(\ref{grad}), we can obtain 
\begin{align*}
 | [\nabla_{\bm \mu} G_t(\bm \mu_t)]_p | 
&\leq  \frac{\lambda}{2} \Vert \bm W_{p, t} \Vert_F^2
 \!+ \frac{1}{|Y_t|}\!\sum_{i \in Y_t} a_t^{(i)} | \hat{\phi}_p(\bm x_t)^\top \! \bm w_{p, t}^{(i)}|  
 \!+ \frac{1}{|\bar Y_t|} \! \sum_{i \in \bar Y_t} b_t^{(i)} | \hat{\phi}_p(\bm x_t)^\top \!\bm w_{p, t}^{(i)} | \\
& \leq \frac{1}{\lambda} + \frac{1}{\lambda} \left( \frac{1}{|Y_t|}\sum_{i \in Y_t} a_t^{(i)} + \frac{1}{|\bar Y_t|} \sum_{i \in \bar Y_t} b_t^{(i)}\right) \leq \frac{3}{\lambda} 
\end{align*}
where Cauchy-Schwarz inequality gives 
$$| \hat{\phi}_p(\bm x_t)^\top \bm w_{p, t}^{(i)}| \leq ||\hat{\phi}_p(\bm x_t)|| \cdot || \bm w_{p, t}^{(i)}|| \leq \frac{1}{\lambda}.$$
Naturally, we obtain that $||\nabla_{\bm \mu} G_t(\bm \mu_t)||_{\infty}\leq \frac{3}{\lambda}$. 

Now, by directly applying the regret bound of EGD (Corollary 5.7 in \cite{hazan2023}), we can prove that OMKC satisfies:
\begin{align} \label{regret1}
\sum_{t=1}^T G_t(\bm\mu_t) - \min_{\bm\mu \in \Delta} \sum_{t=1}^T G_t(\bm\mu) \leq O(\sqrt{T\ln{P}}) .
\end{align}

This sublinear regret bound (\ref{regret1}) reveals that, as $T\rightarrow \infty$,
the average online performance of OMKC matches that of the \emph{optimal fixed} convex combination of its maintained single-kernel multi-label models. 

Together with (\ref{regret2}), these results show that, as $T\rightarrow \infty$, OMKC performs comparably to the \emph{optimal fixed} convex combination of the \emph{best fixed} single-kernel multi-label models chosen in hindsight from different kernel-induced RFF spaces.

\section{Experiments}
\label{sec.experiments}

This section presents three sets of experiments. The first set empirically examines the convergence of OMKC and compares it with OKS-OMC and two methods based on GAM. The second set compares OMKC against state-of-the-art online multi-label classification algorithms and includes an ablation analysis of its label thresholding strategy. The third set presents a sensitivity analysis of the key hyperparameters in OMKC.
All experiments use open benchmark datasets from the Mulan\footnote{\url{http://mulan.sourceforge.net/datasets-mlc.html}} and MLDA\footnote{\url{https://www.uco.es/kdis/mllresources/}} repositories, with their characteristics detailed in Table \ref{data}. 

\begin{table}[!htp]
	\caption{The characteristic of the datasets used in the experiments.} 
	\label{data}
	\footnotesize
	\centering \tabcolsep =2pt
	\begin{threeparttable}
	\begin{tabular}{cccccc}
	\hline
	Datasets     &domains  &\makecell{\#instances \\($T$)}   &\makecell{\#features \\ ($d$)}    &\makecell{\#labels \\ ($L$)}   &cardinality \\
	\hline
       Emotions        &music        &593          &72           &6       &1.868      \\
       Medical         &text        &978          &1449         &45       &1.245     \\
       PlantGO        &biology     &978         &3,091        &12     &1.079        \\
       PlantPseAAC  &biology     &978         &440        &12     &1.079         \\
       Enron            &text         &1,702        &1,001        &53     &3.378       \\
       Image           &image      &2,000	   &294	     &5      &1.236        \\
       Scene           &image      &2,407        &294         &6      &1.074       \\
       Yeast            &biology    &2,417         &103         &14     &4.237      \\
       HumanGO      &biology    &3,106         &9,844     &14     &1.185    \\ 
       HumanPseAAC  &biology   &3,106       &440        &14     &1.185  \\
       Tmc2007-500	    &text        &28,596       &500     &22      &2.220     \\
	\hline  
	\end{tabular}
	\end{threeparttable}
\end{table}

\subsection{Empirical convergence analysis and comparison}
\label{evidence}

In this section, we verify the convergence of OMKC by examining the evolution of its average cumulative loss.
We then establish its advantage through a comparison with OKS-OMC \cite{ZhaiL25} and two variants of the GAM method \cite{JainK17}: GAM ($\mathcal W\rightarrow\mu$) and GAM ($\mu\rightarrow\mathcal W$). 
These two variants are obtained by replacing our proposed optimization strategy with the GAM method, differing only in the update order of the variable blocks in each round. Specifically, GAM ($\mathcal W \rightarrow \bm\mu$) first updates $\mathcal W$ and then $\bm\mu$, whereas GAM ($\bm\mu\rightarrow\mathcal W$) first updates $\bm\mu$ followed by $\mathcal W$. 
For a controlled ablation study focused specifically on the combination coefficient updating strategies of OMKC and OKS-OMC, we adapt the single-kernel models of OKS-OMC by replacing their label thresholding mechanism with the fixed thresholding strategy used in OMKC.  

Experiments are performed on six datasets randomly chosen from Table \ref{data}.
All methods use the same hyperparameter settings on each dataset.
Specifically, for \textsl{Enron}, \textsl{Yeast} and \textsl{Tmc2007-500} datasets, seven polynomial kernels defined on the unit sphere \cite{PenningtonYK15} are used. These kernels take the form $\mathcal{K}_p (\bm x_i, \bm x_j) = (1 - {||\bm x_i -\bm x_j||^2}/{4})^{b}$, where the polynomial order $b \in \{1, 2, \cdots, 7\}$.
For \textsl{Image}, \textsl{Scene} and \textsl{HumanPseAAC} datasets, nine Laplacian kernels of the form
 $\mathcal{K}_p (\bm x_i, \bm x_j) = \exp\bigl(-{||\bm x_i -\bm x_j||_1} /{\sigma}\bigr)$ are used, with $\sigma$ taking values in $\{2^{-2}, 2^{-1},\cdots, 2^{6}\}$. 
In approximating the above two types of kernels, two different values for the number of RFF components (i.e., $D$) are selected for each dataset, based on its size and dimensionality.
On these datasets, other hyperparameters are set as follows: regularization coefficient $\lambda = 10^{-6}$ when the number of instances $T  < 10^4$ and $\lambda = 10^{-8}$ when $T > 10^4$, classifier learning rate $\eta = 0.5$, combination coefficient learning rate $\beta = 0.25$, and global threshold $\gamma = 0$.  
Given that multiple learning on an incoming example can significantly improve online performance \cite{ZhaiW23, ZhaiT2023, ZhaiL25}, all methods are permitted to learn each example $L$ times. 

\begin{figure*}[htp]
	\centering
	\subfloat[]{\includegraphics[width=0.33 \textwidth]{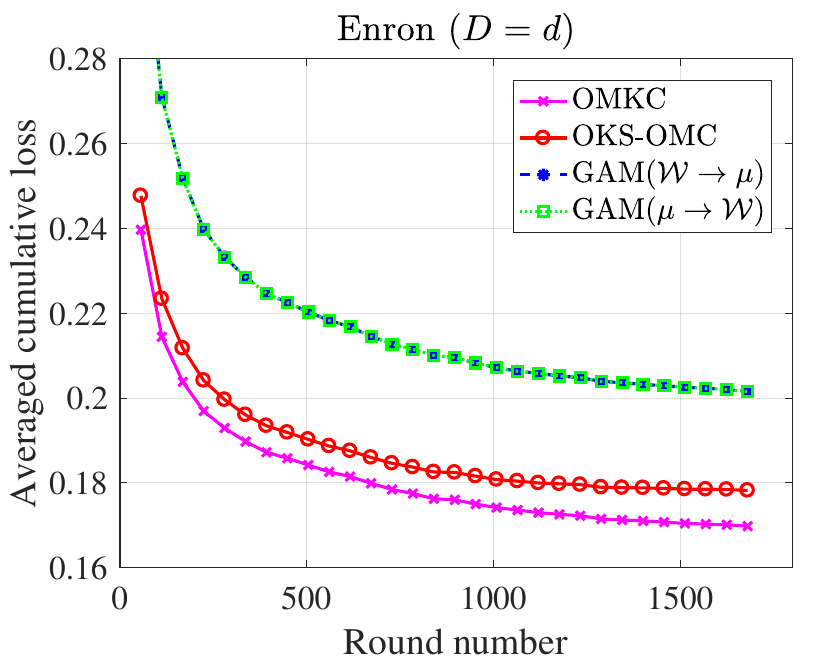}}
	\subfloat[]{\includegraphics[width=0.33 \textwidth]{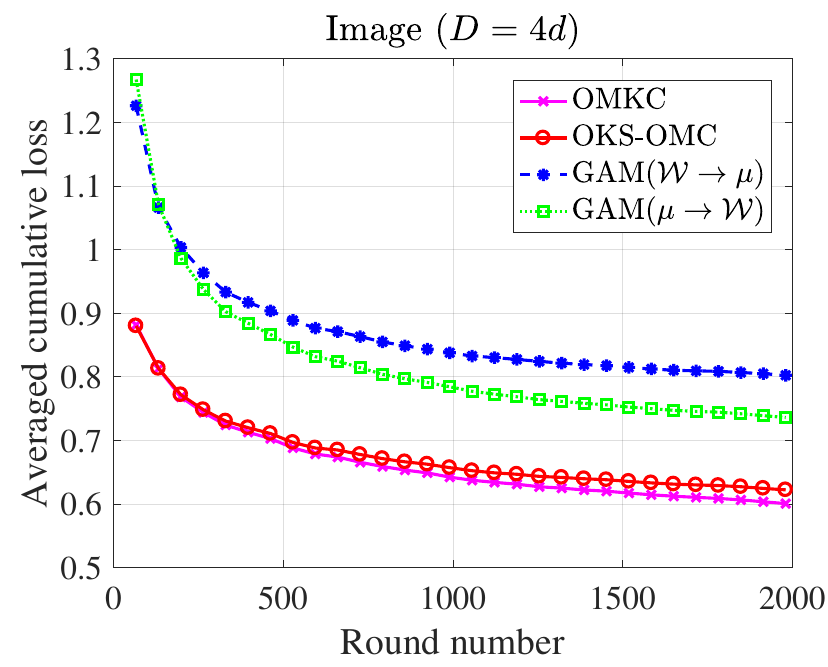}}
	\subfloat[]{\includegraphics[width=0.33 \textwidth]{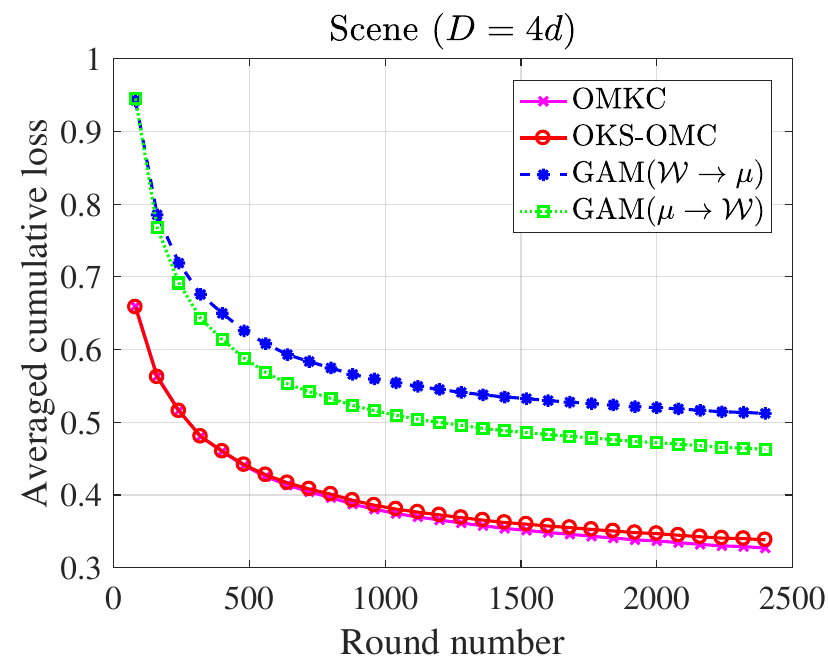}}

	\subfloat[]{\includegraphics[width=0.33 \textwidth]{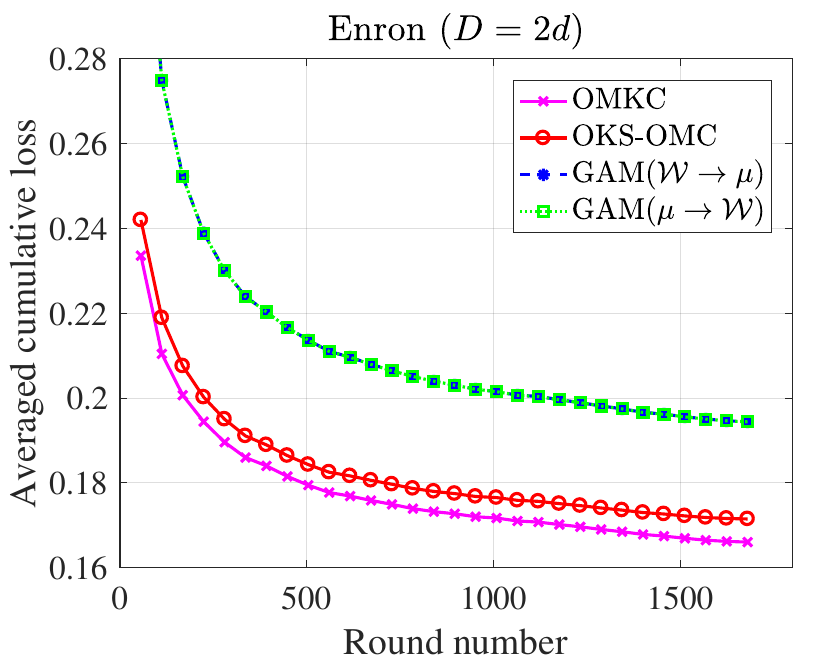}}
	\subfloat[]{\includegraphics[width=0.33 \textwidth]{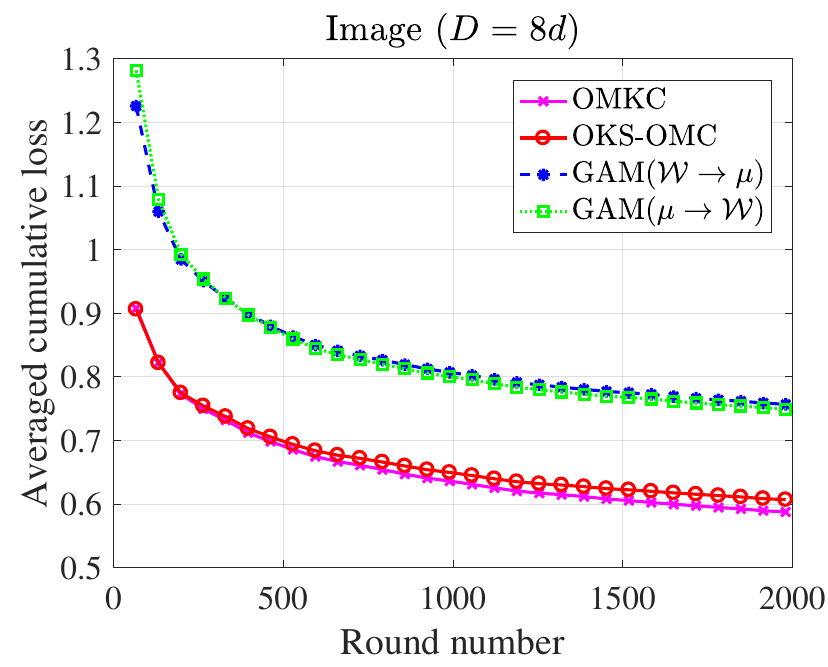}}
	\subfloat[]{\includegraphics[width=0.33 \textwidth]{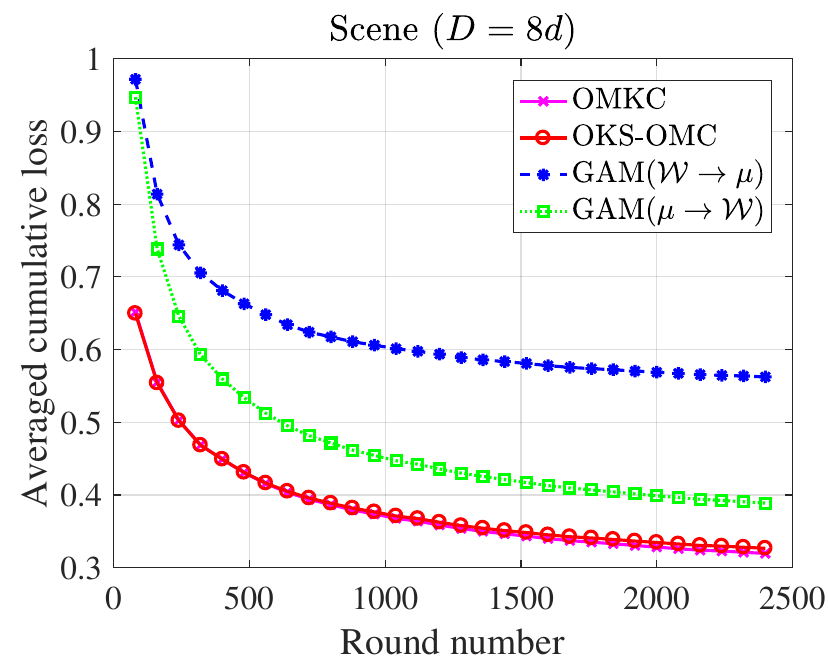}}
	
	\subfloat[]{\includegraphics[width=0.33 \textwidth]{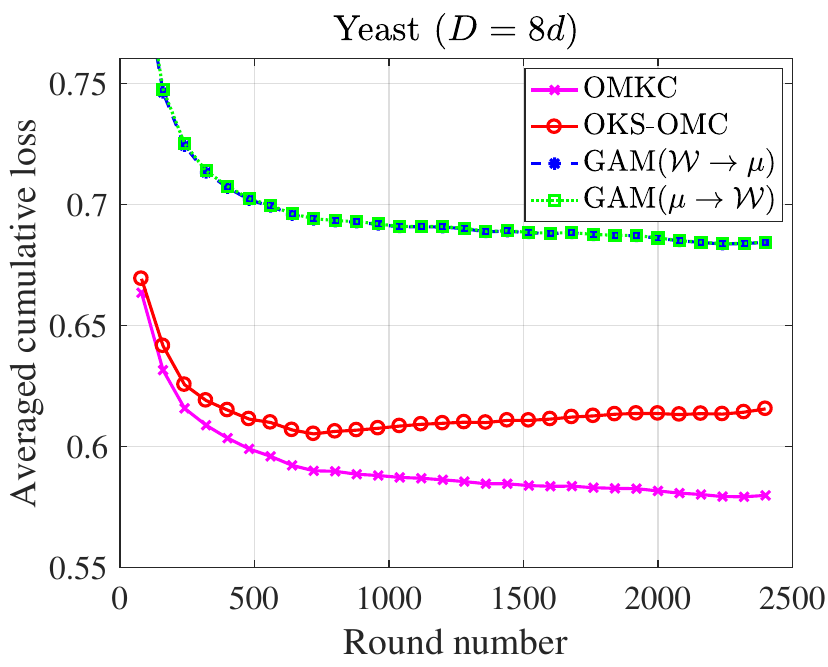}}
	\subfloat[]{\includegraphics[width=0.33 \textwidth]{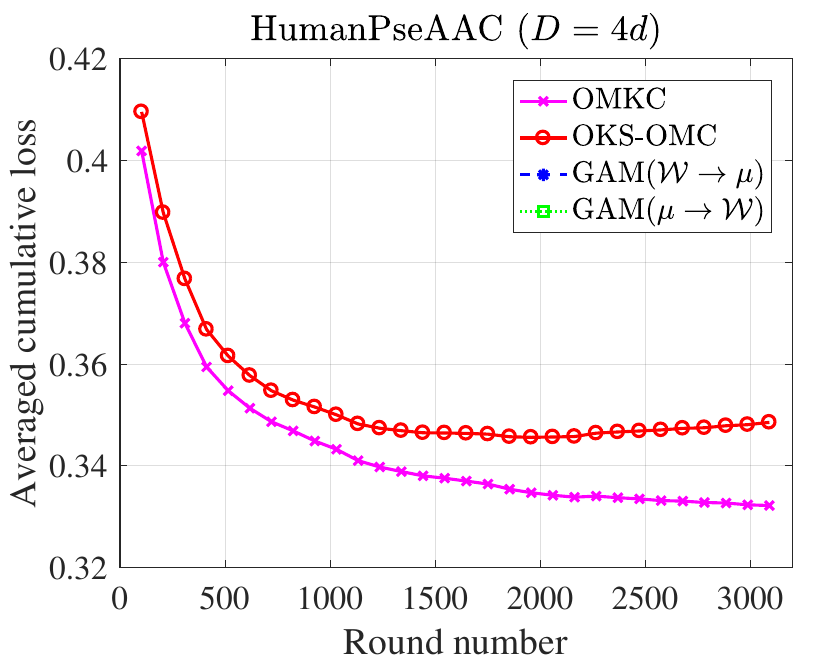}}
	\subfloat[]{\includegraphics[width=0.33 \textwidth]{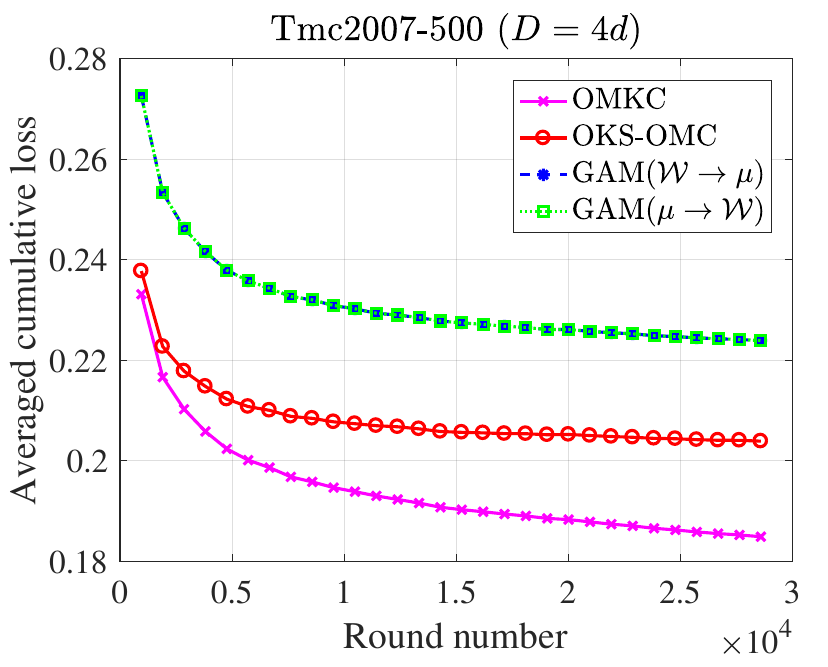}}

	\subfloat[]{\includegraphics[width=0.33 \textwidth]{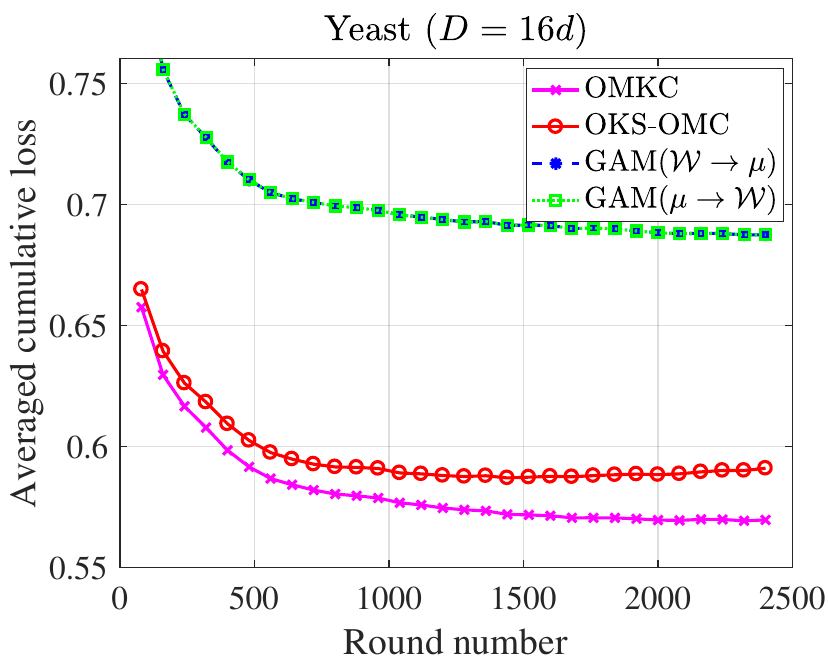}}
	\subfloat[]{\includegraphics[width=0.33 \textwidth]{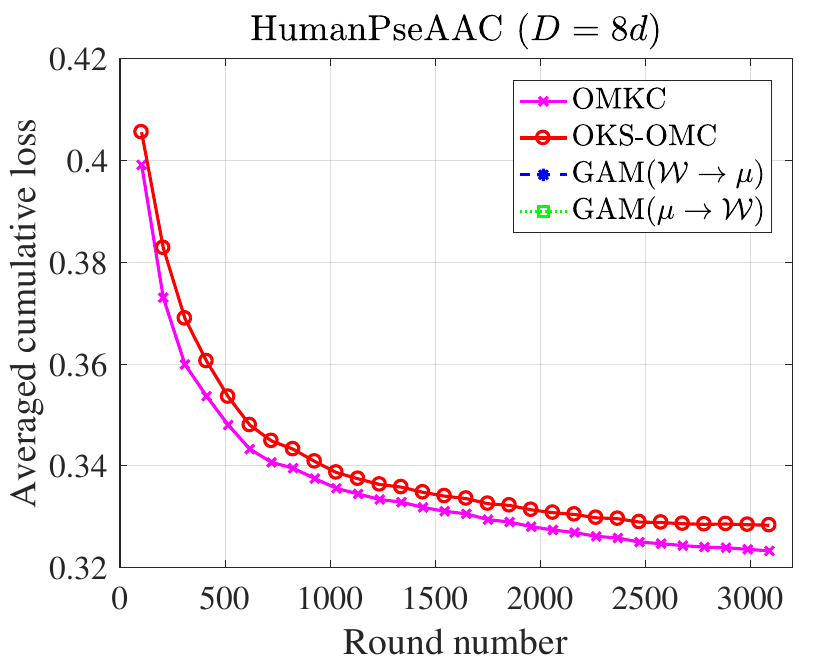}}
	\subfloat[]{\includegraphics[width=0.33 \textwidth]{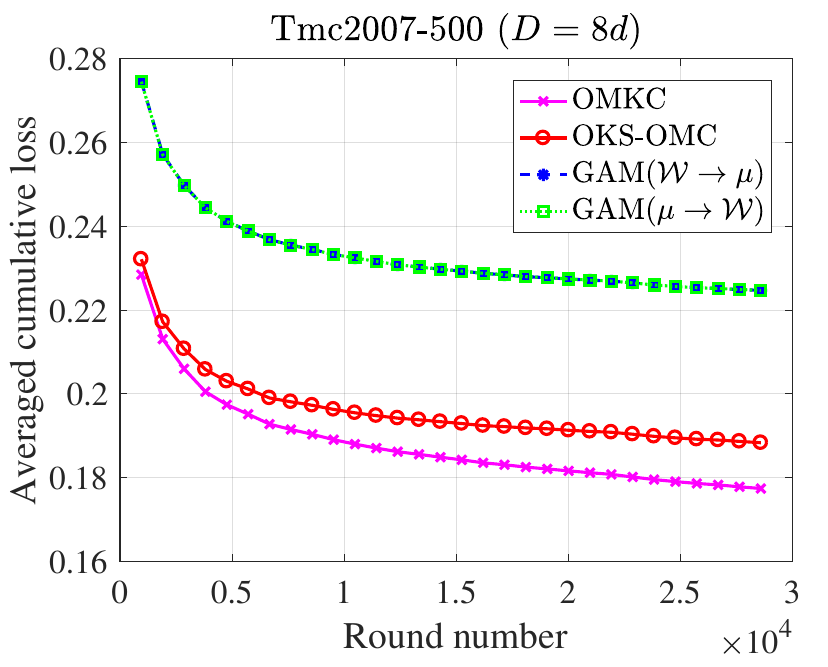}}
	\caption{The evolution of average cumulative loss $Avg(t)$ with the round number $t$ for both OMKC and OKS-OMC. On the \textsl{HumanPseAAC} dataset, the $Avg(t)$ values for the two GAM variants are substantially higher than those of OMKC and OKS-OMC. As a result, their curves fall outside the plotted range of [0.32, 0.42]. }
	\label{acl}
\end{figure*}

We record the average cumulative loss, $Avg(t)$, of all methods, evaluated in a single pass over each dataset. $Avg(t)$ is calculated as
\begin{align*}
Avg(t) = \frac{1}{t*L}\sum_{k=1}^t \sum_{l=1}^L \mathcal{F} (\mathcal{W}_{k,l}, \bm\mu_{k,l}; (\bm x_k, Y_k)).
\end{align*}
where $(\mathcal{W}_{k,l}, \bm\mu_{k,l})$ is the model maintained by each method at round number $k$ and time number $l$.
Fig. \ref{acl} shows the evolution of $Avg(t)$ for all methods as the round number $t$ increases, based on results averaged over 20 independent runs with randomly shuffled examples.

From these figures, it is evident that the average cumulative loss $Avg(t)$ for OMKC exhibits a decreasing trend
with the round number $t$, across all six datasets, indicating its convergence. 
On all datasets, OMKC consistently achieves the lowest $Avg(t)$, followed by OKS-OMC, while the two GAM variants yield the highest values, reflecting their slower convergence.

The difference in $Avg(t)$ between OMKC and OKS-OMC can be attributed to their distinct updating mechanisms for combination coefficients. Specifically, OMKC directly minimizes the objective function $\mathcal{F} (\mathcal{W}, \bm \mu; (\bm x_t, Y_t)) $, whereas OKS-OMC minimizes an upper bound of this function.
In contrast, the performance gap in $Avg(t)$ between OMKC and GAM ($\mu\rightarrow\mathcal W$) arises from their fundamental difference in updating kernel classifiers. GAM ($\mu\rightarrow\mathcal W$) updates classifiers by directly minimizing $\mathcal{F} (\mathcal{W}, \bm \mu_{t+1}; (\bm x_t, Y_t)) $, whereas OMKC minimizes an upper bound of this function. OMKC's design enables it to update classifiers independently of the combination coefficients, unlike GAM ($\mu\rightarrow\mathcal W$), where the updates depend on them. 
This key design of OMKC allows it to quickly focus on promising regions and accelerate convergence to a high-quality solution.

We further observe that a larger kernel approximation error (a smaller $D$) correlates with a greater gap in $Avg(t)$ between OMKC and OKS-OMC. A plausible explanation is that an increased kernel approximation error degrades the accuracy of each single-kernel multi-label model. This degradation likely amplifies the discrepancy between the gradient directions of the true objective function $\mathcal{F} (\mathcal{W}, \bm \mu; (\bm x_t, Y_t))$ and its upper-bound surrogate used by OKS-OMC with respect to $\bm \mu$.

In conclusion, our results show that OMKC's combination coefficient updating strategy outperforms that of OKS-OMC,
and their performance gap widens with larger kernel approximation error.
Furthermore, we show that, in an online setting, the improved alternating optimization strategy of OMKC is superior to the direct alternating optimization strategy of GAM.

\subsection{Comparative experiments}

This section presents a comprehensive evaluation of the following online multi-label classification methods:
\begin{itemize}

\item OSML-ELM \cite{VenkatesanEDPW17} and ELM-OMLL \cite{DuV20}: two methods based on Extreme Learning Machine (ELM). Their number of hidden layer neurons is chosen from $\{100, 200,\cdots, 1000\}$. For ELM-OMLL, its regularization factor is from \{0.001, 0.005, 0.01, 0.05, 0.1, 0.5, 1\}.

\item  ELM-O \cite{ZouHLWH26}: another ELM-based method originally designed for handling noisy labels in data streams.
We evaluate it directly on our datasets without introducing additional label noise. 
Following the authors' recommendations, we tune the weight factor $\beta$ in $\{0.1, 0.2, \cdots, 1\}$, and the regularization factor $\gamma$ in $\{0, 2^{-10}, 2^{-9}, \cdots, 2^0\}$, while keeping all other hyperparameters as default.

\item PAML-I \cite{ZhaiW23} and FFLT \cite{ZhaiT2023}: two single-kernel methods that have been integrated with  
the RFF techniques. For a fair comparison, their best kernel for each dataset is selected from the predefined kernel set used by  OMKC. For PAML-I, its aggressiveness hyperparameter $C$ is searched from $\{2^{-10}, 2^{-9}, \cdots, 2^{0}\}$.
For FFLT, its step-size is set equivalent to that used in OMKC, and each example is also allowed to learn $L$ times per online round.

\item OKS-OMC \cite{ZhaiL25}: an online multi-kernel learning algorithm. To ensure a fair comparison, OKS-OMC is implemented with the same RFF technique and hyperparameter configurations as OMKC on each dataset.

\item OMKC: our method that uses nine Laplacian kernels on the \textsl{Emotions}, \textsl{PlantPseAAC}, \textsl{Image}, \textsl{Scene}, \textsl{HumanPseAAC} datasets, and seven polynomial kernels defined on the unit sphere on the remaining datasets. These kernels follow the same format and hyperparameter ranges described in the previous section. The classifier learning rate $\eta$ is chosen from $\{0.25, 0.5\}$.
The number of RFF components, $D$, is chosen from the set $\{d, 2d, 4d, 8d, 16d, 32d\}$.
To maintain a reasonable computational cost, a smaller $D$ value in this set is used when the original data dimensionality $d$ is larger, and a larger $D$ value is used when $d$ is smaller.  
Other hyperparameters remain unchanged from earlier descriptions.
 
\item OMKC\textsubscript{a}: a variant of our method that replaces the fixed label thresholding mechanism in the single-kernel models with the adaptive thresholding strategy used in OKS-OMC. 
This variant enables an ablation study to determine which label thresholding strategy—when integrated with our combination coefficient update strategy—yields better performance in the high-dimensional RFF-induced feature space.
For this ablation study, OMKC\textsubscript{a} is configured with the same hyperparameters as OMKC.
\end{itemize}

All compared algorithms were implemented using the source code provided by their original authors.
All algorithms are evaluated under the progressive validation framework \cite{BlumKL99}, which is the standard protocol for online learning \cite{Shalev-Shwartz12, Hazan16}. This framework provides a realistic assessment of the sequential prediction performance of an online algorithm, unlike the traditional train-then-test evaluation used for batch methods. 
In this protocol, each dataset is presented as a data stream and fed into each algorithm sequentially.
At each round $t$, the algorithm receives an instance $\bm x_t$, predicts its relevant label set as $\hat Y_t$, and then  receives the true relevant label set $Y_t$. Based on the discrepancy between $\hat Y_t$  and $Y_t$, standard multi-label classification metrics are incrementally updated. At the end of round $t$, the model is updated with the example ($\bm x_t, Y_t$). This iterative test-then-train process continues until the entire stream is processed.

For hyperparameter tuning, we reserve an initial segment of each data stream (5\% for all datasets). Once optimal hyperparameters are identified, this initial segment is also used for model initialization. Each model is then evaluated sequentially on the subsequent examples in this stream.

To ensure a thorough comparison across different performance aspects, 
we employ five example-based metrics—Precision (Psn), Recall (Rcal), F1-measure (F1), Hamming loss (Hl), and Ranking loss (Rl)—along with two label-based metrics, Macro-averaged F1 (MacroF1) and Micro-averaged F1 (MicroF1). These are defined as follows:
\begin{gather*}
\textrm{Psn} = \frac{1}{N} \sum_{i=1}^N \frac{|Y_i \cap \hat Y_i |}{|\hat Y_i|}, \quad
\textrm{Rcal} = \frac{1}{N} \sum_{i=1}^N \frac{|Y_i \cap \hat Y_i |}{|Y_i|}, \quad 
\textrm{F1} = \frac{2*\textrm{Psn}*\textrm{Rcal}}{\textrm{Psn} + \textrm{Rcal}}, \\
\textrm{Hl}  = \frac{1}{N} \sum_{i=1}^N \frac{1}{L} | \hat Y_i  \Delta Y_i |, \quad
\textrm{Rl}  = \frac{1}{N} \!\sum_{i=1}^N \! \frac{\sum_{j \in Y_i, k \in \bar Y_i} \mathbbm{1} [h(\bm x_i, j)  \leq h(\bm x_i, k)]}{|Y_i| * |\bar Y_i|} , \\
\textrm{MacroF1} = \frac{1}{L} \sum_{j=1}^L \frac{2*tp_j}{2*tp_j + fp_j + fn_j}, \
\textrm{MicroF1} = \frac{2* \sum_{j=1}^L tp_j}{\sum_{j=1}^L (2*tp_j + fp_j + fn_j)},
\end{gather*}
where $\Delta$ is the symmetric difference between two sets; $h(\bm x_i, j)$ denotes the predicted score for label $j$ given instance $\bm x_i$; $tp_j, fp_j$ and $fn_j$ denote the counts of \emph{true positives, false positives} and \emph{false negatives} for label $j$, respectively.

To ensure reliable results, each algorithm is executed $20$ times, each with a differently shuffled dataset. 
Average metric values over the $20$ runs are reported in Tables \ref{result1} and \ref{result2}, with the best value on each metric in bold.

\begin{table*}[!htp]
\caption{Online performance metrics [\%] and standard derivation (in brackets) achieved by all compared algorithms on the first six datasets.}
\label{result1}
\centering \tabcolsep = 2pt
\scriptsize
\begin{threeparttable}
\begin{tabular}{llccccccc}
\hline
Dataset & Algorithm  & Psn ($\uparrow$)  & Rcal ($\uparrow$) & F1 ($\uparrow$)  &MacroF1 ($\uparrow$)  & MicroF1 ($\uparrow$)  & Hl ($\downarrow$) & Rl ($\downarrow$) \\
\hline
\multirow{8}{*}{\makecell{Emotions \\ ($D=16d$)}}
&OSML-ELM &60.09 (1.57) &56.97 (1.65) &58.48 (1.50) &59.38 (1.69) &61.28 (1.35) &22.36 (0.69) &19.69 (0.78) \\
&ELM-OMLL &61.77 (0.76) &{74.90 (1.20)} &67.70 (0.76)	&65.98 (0.71) &	66.59 (0.71)  &23.26 (0.48) &18.54 (0.45) \\ 
&ELM-O    &31.89 (0.18) &\textBF{81.61 (0.22)} &45.86 (0.15) &43.82 (0.06) &46.22 (0.16) &61.96 (0.27) &53.28 (0.06) \\
&PAML-I    &62.87 (0.86) &69.17 (1.14)  &65.86 (0.83) &64.53 (0.55) &66.23 (0.50) &21.88 (0.42) &17.96 (0.48) \\
&FFLT        &64.35 (0.77) &70.97 (0.81) &67.49 (0.61) &65.89 (0.59) &67.43 (0.58) &21.15 (0.41) &17.31 (0.43) \\
&OKS-OMC&64.45 (0.76) &69.68 (0.82) &66.96 (0.50) &65.35 (0.61) &67.03 (0.52) &21.11 (0.38) &17.09 (0.53) \\
&OMKC\textsubscript{a}  &\textBF{64.84 (0.93)} &69.13 (0.73) &66.91 (0.56) &65.35 (0.44) &67.11 (0.41) &\textBF{20.91 (0.35)} &\textBF{17.07 (0.42)} \\
&OMKC     &{64.66 (0.65)} &71.36 (0.80) &\textBF{67.84 (0.50)} &\textBF{66.00 (0.46)} &\textBF{67.60 (0.42)} &21.12 (0.31) &17.21 (0.49) \\
\hline
\multirow{8}{*}{\makecell{Medical \\ ($D=4d$)}}
&OSML-ELM &54.45 (2.24) &51.16 (2.05)  &52.75  (2.13)  &14.24 (1.75)  &63.08 (1.44)  &1.65 (0.05)  &4.74 (0.49) \\
&ELM-OMLL &56.52 (1.28)  &\textBF{87.04 (0.68)} &68.53 (0.99) &\textBF{33.68 (1.11)} &59.90 (0.87) &3.18 (0.11) &4.80 (0.30)  \\
&ELM-O  &27.02 (0.00) &21.21 (0.00) &23.76 (0.00) &0.95 (0.00) &24.06 (0.01) &3.79 (0.00) &24.96 (0.17) \\
&PAML-I      &53.75 (0.80)  &50.02 (0.74) &51.82 (0.75)  &17.58 (1.15) &62.45 (0.53) &1.63 (0.02) &3.87 (0.13) \\
&FFLT         &59.90 (0.76)  &60.83 (0.73) &60.36 (0.69) &20.60 (1.13) &66.28 (0.57) &1.69 (0.04) &4.44 (0.17) \\
&OKS-OMC&64.80 (0.61)  &66.76 (0.61) &65.77 (0.56) &24.56 (1.01)  &70.80 (0.48)  &1.50 (0.03) &4.18 (0.16)  \\
&OMKC\textsubscript{a} &65.10 (0.65) &67.18 (0.67) &66.12 (0.62) &24.95 (1.00) &71.06 (0.52) &\textBF{1.49 (0.03)} &4.16 (0.17) \\
&OMKC     &\textBF{67.17  (0.77)}  &70.88 (0.62) &\textBF{68.98 (0.66)} &26.27 (1.11)  &\textBF{72.09 (0.55)}  &\textBF{1.49 (0.03)} &\textBF{3.69 (0.19)}  \\
\hline
\multirow{8}{*}{\makecell{PlantGO \\ ($D = d$)}}
&OSML-ELM &62.25 (2.03) &62.62 (2.29) &62.43 (2.15) &38.95 (4.82) &70.47 (1.57) &\textBF{4.61 (0.18)} &5.23 (0.48)  \\
&ELM-OMLL &61.57 (1.27)   &\textBF{80.68 (1.13)} & 69.84 (1.12) &58.50 (1.17) &64.58 (0.92) &7.90 (0.25) &7.99 (0.53) \\ 
&ELM-O      &28.11 (1.40) &28.71 (3.45) &28.23 (0.26) &3.80 (0.51) &27.36 (0.33) &12.90 (1.27) &27.46 (0.60) \\
&PAML-I       &62.98 (1.05)  &63.67 (1.58) &63.32 (1.28) &44.27 (1.74) &69.40 (0.85) &4.92 (0.15) &5.86 (0.28) \\
&FFLT          &67.43 (0.75)  &74.93 (0.84) &70.98 (0.74) &56.80 (1.01) &72.26 (0.78) &5.08 (0.17) &5.54 (0.20) \\
&OKS-OMC &69.41 (0.62)  &77.34 (0.67) &73.16 (0.59) &60.13 (0.97)  &73.96 (0.69) &4.83 (0.16) &5.19 (0.23) \\
&OMKC\textsubscript{a} &\textBF{69.59 (0.69)} &77.42 (0.56) &73.29 (0.58) &59.98 (0.77) &\textBF{74.09 (0.68)} &{4.80 (0.15)} &5.11 (0.24) \\
&OMKC       &{69.43 (0.76)}  &78.37 (0.51) &\textBF{73.63 (0.60)} &\textBF{60.97 (1.31)} &73.85 (0.71)  &4.92 (0.16) &\textBF{5.00 (0.22)} \\
\hline
\multirow{8}{*}{\makecell{PlantPseAAC \\ (D = 8d)}}
&OSML-ELM &12.28 (1.42) &12.15 (1.44)   &12.21 (1.43)  &7.39 (1.19)	&18.88 (1.97)  &\textBF{8.96 (0.11)}  &21.03 (0.54)  \\
&ELM-OMLL &26.53 (1.11)  &40.83 (1.60)  &32.16 (1.28) &\textBF{20.16 (0.98)}	&31.74 (1.19)  &	15.50 (0.34) &29.30	(0.76)  \\
&ELM-O      &23.86 (0.40) &44.62 (0.29) &31.09 (0.41) &6.37 (0.10) &31.02 (0.53) &17.69 (0.14) &27.07 (0.50) \\
&PAML-I      &27.27 (1.29)  &44.84 (3.15)  &33.88 (1.59) &17.85 (1.36)  &34.29 (1.08)  &15.17 (1.01)  &21.50 (0.57) \\
&FFLT          &31.66 (0.60) &47.28 (1.23)  &37.92 (0.65) &14.94 (0.72) &37.11 (0.61)  &{14.08 (0.29)}  &\textBF{20.00 (0.39)} \\
&OKS-OMC  &\textBF{31.74 (0.50)} &47.08 (1.14)  &37.91  (0.57) &14.42 (0.66) &36.81 (0.52) &14.19 (0.28)  &\textBF{20.00 (0.37)} \\
&OMKC\textsubscript{a} &31.50 (0.60) &46.65 (1.14) &37.60 (0.69) &15.75 (0.78) &36.53 (0.66) &14.25 (0.27) &20.02 (0.35) \\
&OMKC      &31.39 (0.64)  &\textBF{48.72 (1.13)}  &\textBF{38.18 (0.69)}  &17.10 (0.80)  &\textBF{37.24 (0.68)} &14.45 (0.29) &20.02 (0.31) \\
\hline
\multirow{8}{*}{\makecell{Enron  \\ (D = 2d)}}
&OSML-ELM &\textBF{64.38 (0.97)} &42.67 (0.99) &51.32 (0.93)	 &9.67 (0.58)	&51.29 (0.65)  &\textBF{4.91 (0.04)} &10.63 (0.18) \\
&ELM-OMLL &54.88 (0.59) &51.65 (0.58)  &53.21 (0.50)	 &13.52 (0.59)  &50.96 (0.46) &6.05 (0.07) &15.12 (0.34) \\
&ELM-O    &43.64 (0.03) &55.46 (0.01) &48.84 (0.02) &4.55 (0.00) &47.35 (0.03) &7.33 (0.00) &12.55 (0.10) \\
&PAML-I     &62.44 (0.77)  &43.65 (0.66)  &51.38 (0.64) &13.32 (0.38) &50.87 (0.51) &{5.06 (0.04)}   &11.39 (0.21) \\
&FFLT          &60.76 (0.60) &58.84 (0.37) &59.78 (0.41) &\textBF{19.64 (0.47)} &57.94 (0.28) &5.27 (0.04)   &10.37 (0.20)\\
&OKS-OMC  &61.79 (0.52) &57.35 (0.35) &59.49 (0.36) &18.02 (0.52) &57.29 (0.27) &5.22 (0.04)   &9.06 (0.13) \\
&OMKC\textsubscript{a} &{63.23 (0.47)} &58.03 (0.48) &60.52 (0.39) &18.09 (0.48) &58.13 (0.23) &5.09 (0.03) &\textBF{8.63 (0.11)} \\
&OMKC        &62.09 (0.46)  &\textBF{59.31 (0.39)}  &\textBF{60.67 (0.36)} &19.51 (0.50)  &\textBF{58.62 (0.29)}  &5.15 (0.04)   &9.88 (0.19) \\
\hline
\multirow{8}{*}{\makecell{Image \\ (D = 16d)}}
&OSML-ELM &51.30 (0.84) &49.83 (0.95) &50.55 (0.85)  &55.27 (0.65)	&55.22 (0.59)	&18.95 (0.21)  &19.17 (0.41) \\
&ELM-OMLL &59.07 (0.60) &{72.61 (0.66)} &65.15 (0.56)  &61.69 (0.49)	&61.34 (0.51)	&21.80 (0.34) &18.51 (0.38) \\
&ELM-O    &24.56 (0.01) &\textBF{88.06 (0.26)} &38.40 (0.04) &38.21 (0.02) &38.40 (0.04) &69.33 (0.04) &51.14 (0.46) \\
&PAML-I     &60.01 (0.69) &58.42 (0.71) &59.20 (0.69) &61.84 (0.45) &61.33 (0.48) &\textBF{17.36 (0.20)} &16.90 (0.19) \\
&FFLT   	    &63.57 (0.60) &69.56 (0.73) &\textBF{66.43 (0.62)} &64.78 (0.53) &64.10 (0.52) &18.42 (0.26) &15.84 (0.25) \\
&OKS-OMC &63.09 (0.66) &69.12 (0.92) &65.97 (0.70) &64.50 (0.59) &63.84 (0.54) &18.49 (0.30) &15.67 (0.21) \\
&OMKC\textsubscript{a} &63.90 (0.56) &67.80 (0.71) &65.79 (0.60) &64.79 (0.57) &64.11 (0.57) &17.88 (0.27) &\textBF{15.46 (0.21)} \\
&OMKC      &\textBF{64.32 (0.62)}  &68.21 (0.68)  &66.21 (0.59)  &\textBF{65.17 (0.55)}  &\textBF{64.51 (0.55)} &17.73 (0.28)  &\textBF{15.46 (0.23)} \\
\hline
\end{tabular}
\begin{tablenotes}
\footnotesize
 \item [1] $\uparrow$ indicates the higher the better, while $\downarrow$ means the lower the better.
\end{tablenotes}
\end{threeparttable}
\end{table*}

\begin{table*}[!htp]
\caption{Online performance metrics [\%] and standard derivation (in brackets) achieved by all compared algorithms on the last five datasets.}
\label{result2}
\centering \tabcolsep = 2pt
\scriptsize
\begin{threeparttable}
\begin{tabular}{llccccccc}
\hline
Dataset & Algorithm  & Psn ($\uparrow$)  & Rcal ($\uparrow$) & F1 ($\uparrow$)  &MacroF1 ($\uparrow$)  & MicroF1 ($\uparrow$)  & Hl ($\downarrow$) & Rl ($\downarrow$) \\
\hline
\multirow{8}{*}{\makecell{Scene \\ (D = 16d)}}
&OSML-ELM &59.08 (0.55) &59.49 (0.64) &59.28 (0.57) &67.52 (0.52)	&66.85 (0.50)	&10.35 (0.14)  &9.77 (0.25) \\
&ELM-OMLL &63.52 (0.53)  &	73.81 (0.79) &68.28 (0.60) &68.78 (0.45) &67.21 (0.44)   &12.69 (0.18)   &11.67 (0.30) \\
&ELM-O      &18.10 (0.00) &\textBF{91.93 (0.25)} &30.24 (0.02) &30.06 (0.03) &30.21 (0.02) &76.17 (0.17) &49.96 (0.08) \\
&PAML-I      &67.76 (0.49) &68.52 (0.55) &68.13 (0.51) &74.08 (0.41) &73.30 (0.41) &\textBF{8.75 (0.13)} &7.36 (0.17)\\
&FFLT          &72.23 (0.46) &{79.92 (0.40)} &75.88 (0.40) &76.46 (0.37) &74.89 (0.36) &9.42 (0.14) &7.08 (0.17)\\
&OKS-OMC  &71.60 (0.49) &79.74 (0.41) &75.45 (0.38) &76.08 (0.35) &74.51 (0.36) &9.59 (0.16) & 6.92 (0.18) \\
&OMKC\textsubscript{a} &72.35 (0.37) &79.05 (0.36) &75.55 (0.30) &76.73 (0.28) &75.28 (0.30) &9.12 (0.14) &\textBF{6.71 (0.16)} \\
&OMKC       &\textBF{72.93 (0.32)} &79.69 (0.34) &\textBF{76.16 (0.31)} &\textBF{77.06 (0.34)} &\textBF{75.60 (0.33)} &9.04 (0.13) &{6.86 (0.16)} \\
\hline
\multirow{8}{*}{\makecell{Yeast \\ (D = 32d)}}
&OSML-ELM &66.65 (0.35) &57.58 (0.36)  &61.79 (0.28) &36.98 (0.46)  &61.32 (0.17) &21.69 (0.12) &19.62 (0.15) \\
&ELM-OMLL &59.68 (0.27) &\textBF{74.01 (0.32)}  &66.07 (0.23) &41.78 (0.47)	&64.73 (0.19) &24.04 (0.16) &18.20 (0.15) \\
&ELM-O    &51.05 (0.06) &69.67 (0.29) &58.92 (0.07) &29.22 (0.20) &58.36 (0.05) &29.53 (0.03) &21.32 (0.03) \\
&PAML-I      &63.04 (0.38) &70.98 (0.81) &66.77 (0.33) &\textBF{43.13 (0.41)} &65.18 (0.24) &22.61 (0.18) &18.39 (0.22) \\
&FFLT          &64.88 (0.25) &68.17 (0.27) &66.49 (0.20) &42.58 (0.23) &65.59 (0.14) &21.40 (0.10) &18.46 (0.16)\\
&OKS-OMC  &66.72 (0.29) &67.16 (0.27) &66.94 (0.22) &41.53 (0.35) &66.08 (0.19) &20.59 (0.15) &{17.29 (0.13)} \\
&OMKC\textsubscript{a}  &67.17 (0.28) &67.29 (0.26) &\textBF{67.23 (0.20)} &41.62 (0.38) &\textBF{66.41 (0.22)} &20.32 (0.14) &\textBF{16.98 (0.14)} \\
&OMKC       &\textBF{67.18 (0.28)} &66.93 (0.26) &{67.05 (0.21)} &41.51 (0.37) &66.32 (0.20) &\textBF{20.31 (0.13)} &17.53 (0.16) \\
\hline
\multirow{8}{*}{\makecell{HumanGo \\ ($D = d$)}}  
&OSML-ELM & 71.19 (0.69) &73.62 (0.86)  &72.39 (0.75)  &47.41 (1.79) &73.86 (0.52)  &4.21 (0.07) &3.28 (0.12)  \\ 
 &ELM-OMLL &63.51 (0.57) &80.16 (0.49) &70.87 (0.52) &56.13 (1.23) &65.91 (0.50) &6.81 (0.12) &7.31 (0.33) \\
 &ELM-O     &29.49 (0.03) &48.02 (0.04) &36.54 (0.03) &6.49 (0.01) &37.05 (0.03) &14.32 (0.00) &20.46 (0.35) \\
 &PAML-I     &69.70 (0.36) &70.53 (0.44) &70.11 (0.39) &48.90 (1.01) &72.90 (0.32) &4.22 (0.05) &3.31 (0.10) \\
 &FFLT         &74.89 (0.31) &\textBF{84.53 (0.27)} &\textBF{79.42 (0.25)} &\textBF{61.83 (0.79)} &76.87 (0.23) &4.18 (0.05) &2.98 (0.08) \\ 
 &OKS-OMC &74.73 (0.33) &83.23 (0.36) &78.75 (0.31) &60.51 (0.68) &76.61 (0.31)  &4.17 (0.06) &2.98 (0.08) \\
 &OMKC\textsubscript{a} &74.82 (0.37) &83.04 (0.41) &78.72 (0.35) &60.08 (0.86) &76.75 (0.28) &\textBF{4.13 (0.05)} &2.98 (0.07) \\
 &OMKC       &\textBF{74.96 (0.29)} &84.29 (0.32) &79.35 (0.26) &61.33 (0.72) &\textBF{76.93 (0.20)} &4.16 (0.04) &\textBF{2.94 (0.09)} \\
 \hline
\multirow{8}{*}{\makecell{HumanPseAAC \\ ($D = 8d$)}}  
&OSML-ELM  &19.74 (0.90)  &18.84 (0.97)  &	19.27 (0.92)  &9.29 (0.52)	 &26.48 (1.06)	&\textBF{8.48 (0.07)}  &17.95 (0.33) \\
 &ELM-OMLL &33.84 (0.72) &43.38 (0.78) &38.02 (0.71)  &16.17 (0.47) &39.29 (0.64)  &11.04 (0.13) &	20.42 (0.38) \\
 &ELM-O     &29.83 (0.07) &47.89 (0.43) &36.76 (0.08) &6.73 (0.05) &37.06 (0.07) &14.20 (0.07) &19.68 (0.01) \\
 &PAML-I      &26.88 (0.82) &27.14 (1.04) &27.01 (0.91) &12.44 (0.54) &33.22 (0.75) &{8.90 (0.08)}    &17.91 (0.23)  \\
 &FFLT          &36.53 (0.29) &57.94 (0.52) &44.81 (0.29) &19.13 (0.31) &44.71 (0.27) &12.03 (0.08)   &15.74 (0.14) \\
 &OKS-OMC  &\textBF{36.82 (0.43)} &57.49 (0.59) &44.89 (0.42) &19.04 (0.37) &44.38 (0.34) &12.10 (0.11)  &\textBF{15.14 (0.14)} \\
 &OMKC\textsubscript{a} &36.32 (0.44) &56.68 (0.64) &44.27 (0.48) &18.97 (0.47) &43.37 (0.34) &12.42 (0.10) &15.46 (0.17) \\
 &OMKC        &36.71 (0.32)  &\textBF{58.49 (0.54)}  &\textBF{45.11 (0.34)}  &\textBF{19.46 (0.33)} &\textBF{44.85 (0.29)} &12.06 (0.09)  &15.66 (0.17) \\
 \hline
\multirow{8}{*}{\makecell{Tmc2007-500 \\ ($D = 8d$)}}
&OSML-ELM &\textBF{73.07 (0.10)} &63.98 (0.21) &68.22 (0.15) &48.07 (0.39) &66.22 (0.12) &6.10 (0.02) &5.32 (0.03) \\
&ELM-OMLL &68.47 (0.17) &75.12 (0.19) &71.64 (0.12) &47.84 (0.50) &68.80 (0.10) &6.48 (0.02) &5.30 (0.04) \\
&ELM-O       &39.95 (0.02) &57.37 (0.01) &47.10 (0.02) &7.61 (0.00) &45.93 (0.02) &12.83 (0.00) &16.70 (0.01) \\
&PAML-I      &70.28 (0.15) &69.43 (0.14) &69.85 (0.13) &59.19 (0.24) &67.72 (0.12) &6.33 (0.02) &4.82 (0.03) \\
&FFLT         &68.85 (0.12) &81.41 (0.09) &74.60 (0.10) &63.46 (0.18) &71.48 (0.08) &6.28 (0.02) &4.46 (0.02) \\
&OKS-OMC &70.46 (0.08) &79.18 (0.18) &74.57 (0.10) &62.41 (0.21) &71.28 (0.09) &6.14 (0.01)   &3.89 (0.02) \\
&OMKC\textsubscript{a} &{72.25 (0.12)} &80.35 (0.12) &76.08 (0.10) &63.92 (0.21) &72.93 (0.08) &5.74 (0.02) &\textBF{3.56 (0.02)} \\
&OMKC       &{71.47 (0.09)} &\textBF{82.23 (0.09)} &\textBF{76.48 (0.07)} &\textBF{65.91 (0.20)} &\textBF{73.69 (0.07)} &\textBF{5.68 (0.02)} &3.95 (0.02)\\ 
\hline
\end{tabular}
\begin{tablenotes}
\footnotesize
 \item [1] $\uparrow$ indicates the higher the better, while $\downarrow$ means the lower the better.
\end{tablenotes}
\end{threeparttable}
\end{table*}

\subsubsection{An ablation study on the label thresholding strategy}

Based on the results of Tables \ref{result1} and \ref{result2}, we first conduct a statistical comparison between OMKC and OMKC\textsubscript{a} using Wilcoxon signed-rank tests \cite{Demsar06}. This comparison aims to validate our design choice of employing a fixed label thresholding strategy in the single-kernel multi-label model of OMKC.
Table \ref{wilcoxon} reports the Wilcoxon signed-rank test statistics $R^+$ and $R^-$, where $R^+$ denotes the sum of ranks for the datasets on which OMKC outperforms OMKC\textsubscript{a}, and $R^-$ denotes the sum of ranks for the opposite. For a confidence level of $\alpha=0.05$ and $N=11$ datasets, the null hypothesis can be rejected if $\min\{R^+, R^-\} \leq 10$.

\begin{table}[!htp]
\caption{Rank statistics of Wilcoxon signed-rank tests on all evaluation metrics.}
\label{wilcoxon}
\centering 
\tabcolsep = 2pt
\footnotesize
\begin{tabular}{cccccccc}
\hline
Statistics      &Psn    &Rcal   &F1       &MacroF1   &MicroF1   &Hl    &Rl  \\
\hline      
$R^+$          &36	 &65     &64       &65           &62 	      &32 	&17.5 \\
$R^-$          &30 	 &1 	    &2 	   &1 	     &4 	      &34 	&48.5 \\
\hline  
min\{$R^+, R^-$\} &30    &1     &2         &1          &4         &32     &17.5 \\
\hline
\end{tabular}
\end{table}

As shown in Table \ref{wilcoxon}, on Rcal, F1, MacroF1 and MicroF1, the obtained statistics allow us to reject the null hypothesis, indicating that OMKC performs significantly better than OMKC\textsubscript{a} on these metrics. For all other metrics, the null hypothesis cannot be rejected, suggesting comparable performance.

To sum up, OMKC shows superior overall performance compared to OMKC\textsubscript{a}. This justifies our choice of employing a fixed label thresholding strategy—rather than an adaptive one—in the high-dimensional RFF-induced feature space.

\subsubsection{Results analysis}

According to Tables \ref{result1} and \ref{result2}, we first compute the average rank across all 11 datasets in each metric for each compared algorithm, as presented in Table \ref{ranks}.
These metric-wise ranks are then averaged over all metrics to obtain an overall rank for each algorithm, reported in the last column of Table \ref{ranks}. 

\begin{table}[!htp]
\caption{Average ranks per metric and overall ranks over all metrics.}
\label{ranks}
\centering 
\tabcolsep = 2pt
\footnotesize
\begin{tabular}{ccccccccc}
\hline
Algorithm          & Psn    & Rcal   & F1    &MacroF1    & MicroF1 & Hl  & Rl  &Rank \\
\hline      
OSML-ELM	&4.64  &6.45   &5.91  &5.91     &5.45 &3.00   &4.91   &5.18  \\ 
ELM-OMLL	&5.27  &3.27 	&3.91  &3.45    &4.91  &5.73 	&5.73  &4.61  \\
ELM-O	      &6.73   &4.27 	&6.73  &7.00 	&6.73  &7.00 	&6.73  &6.45 \\
PAML-I	      &4.36   &5.27 	&5.00  &4.09 	&4.91  &3.32 	&4.18  &4.45 \\
FFLT	             &3.09   &2.73 	&2.45  &2.45 	&2.27  &3.64 	&3.18  &2.83 \\
OKS-OMC     &2.36  &3.45 	&2.82  &3.27 	&2.64  &3.00 	&1.64  &2.74 \\
OMKC	      &1.55  &2.55 	&1.18   &1.82 	&1.09  &2.32 	&1.64  &1.73  \\
\hline
\end{tabular}
\end{table}

From these results, we observe that OMKC achieves the highest overall rank, owing to its highest average rank in each metric. It is followed by OKS-OMC and FFLT, whose metric-wise ranks are mostly second or third.
Next are PAML-I and ELM-OMLL, while OSML-ELM and ELM-O have the lowest overall ranks, suggesting their inferior overall performance.   

Further, we notice that the three ELM-based methods are inferior to the kernel-based methods in terms of overall performance. The inferiority probably stems from the inherent design mechanism of ELM. 
Specifically, ELM is a feedforward neural network with a single hidden layer. Its input weights and biases of hidden nodes are randomly assigned once and never updated; only the weights of output nodes are incrementally learnt. This frozen hidden layer results in a nonlinear feature mapping that is fixed and lacks adaptability to the data. In contrast, the kernel-based methods can discover more effective feature mappings by searching for appropriate kernels. 
 
Among the kernel-based methods, the single-kernel method PAML-I performs the worst. This is due to its update rule, which updates only two label scoring predictors per round, resulting in insufficient updates for the remaining predictors. In contrast, the single-kernel FFLT updates all label scoring predictors per round, leading to better performance. However, both single-kernel methods are outperformed by the multi-kernel methods OKS-OMC and OMKC.

Next, we focus on comparing OMKC with its most relevant and competitive baselines: FFLT and OKS-OMC.
In order to statistically compare them, we conducted Friedman tests followed by post-hoc Bonferroni-Dunn tests \cite{Demsar06}. Table \ref{statistic} lists the average ranks of the three algorithms across 11 datasets and the corresponding Friedman statistic $F_F$ for each metric.

\begin{table}[!htp]
\caption{The average ranks of FFLT, OKS-OMC and OMKC and the Friedman statistic $F_F$ in terms of each metric.}
\label{statistic}
\centering 
\tabcolsep = 2pt
\footnotesize
\begin{tabular}{cccccccc}
\hline
Algorithm  & Psn    & Rcal   & F1    &MacroF1    & MicroF1    & Hl  & Rl  \\
\hline      
FFLT          &2.636 	 &1.818 	&2.182 	&1.909 	&2.273 	&2.455 	&2.818 \\
OKS-OMC  &2.091 	 &2.545 	&2.636 	&2.727 	&2.636 	&2.091 	&1.545 \\
OMKC        &1.273 	 &1.636 	&1.182 	&1.364 	&1.091 	&1.455 	&1.636 \\
\hline
$F_F$ statistic  &8.906 	&3.011 	&12.407 	&8.906 	&18.810 	&3.444 	&10.167 \\
\hline
\end{tabular}
\end{table}

Given three algorithms and 11 datasets,  the $F_F$ statistic follows an F-distribution with 2 and 20 degrees of freedom.  
The critical value of F(2, 20) for a significance level $\alpha = 0.05$ is 3.493.
As shown in Table \ref{statistic}, the $F_F$ values for Psn, F1, MacroF1, MicroF1 and Rl exceed this critical value. Therefore, for these metrics, we can reject the null hypothesis that these algorithms perform equivalently.
In contrast, for Rcal and Hl, the $F_F$ values are below the critical value. This indicates no statistically significant difference on these metrics.

\begin{figure*}[htp]
	\centering
	\subfloat[]{\includegraphics[width=0.33 \textwidth]{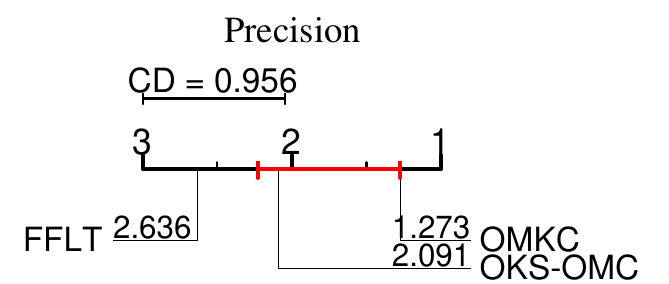}}
	\subfloat[]{\includegraphics[width=0.33 \textwidth]{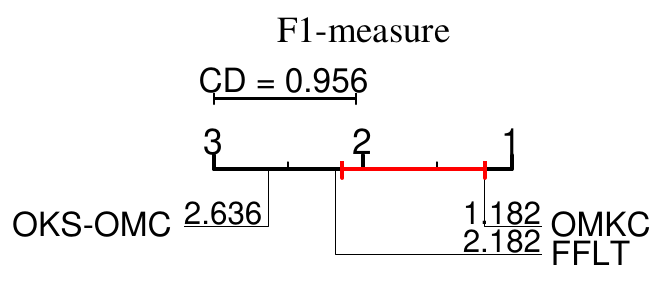}}
	\subfloat[]{\includegraphics[width=0.33 \textwidth]{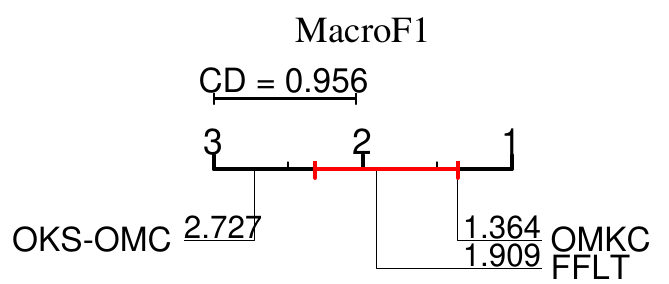}}
	
	\subfloat[]{\includegraphics[width=0.33 \textwidth]{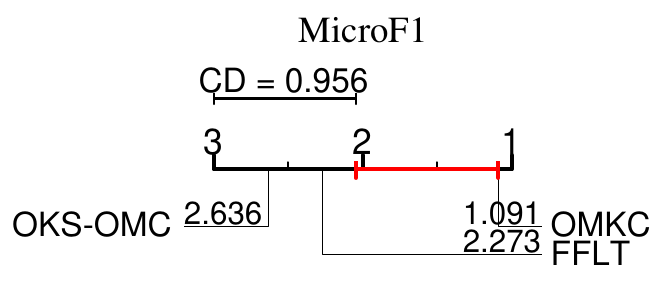}}
	\subfloat[]{\includegraphics[width=0.33 \textwidth]{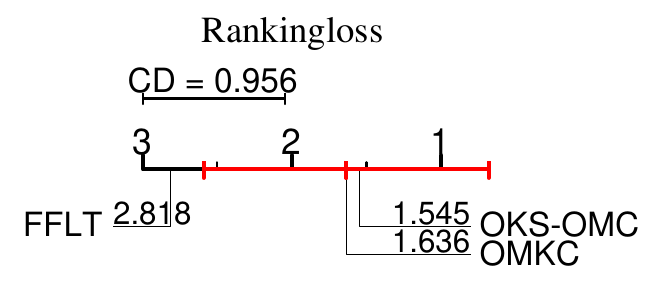}}
	\caption{Statistical comparison of OMKC with two baseline methods using the Bonferroni–Dunn test.}
	\label{CD}
\end{figure*}

Next, on the metrics showing significant Friedman test results, we applied the Bonferroni-Dunn tests \cite{Demsar06} to compare OMKC against the other two algorithms. The critical difference (CD) for $\alpha = 0.05$ is 0.956.
A difference in average ranks that exceeds this CD indicates a statistically significant difference. 
Figure \ref{CD} presents these comparisons visually using CD diagrams. The diagrams plot the average ranks of algorithms on the top axis, and the interval of one CD to either side of OMKC's average rank is highlighted in red.
An algorithm is deemed statistically different from OMKC if its rank falls outside the red interval.

According to Table \ref{statistic} and Figure \ref{CD}, we can draw the following observations:
\begin{itemize}
\item OMKC vs FFLT: 
OMKC demonstrates statistically significant superiority over FFLT on four key metrics: Psn, F1, MicroF1, and Rl. While OMKC also achieves higher average ranks on Rcal, MacroF1, and Hl, these differences are not statistically significant, indicating comparable performance between the two algorithms on these metrics. 
In summary, the evidence confirms that OMKC delivers superior overall performance.
This performance difference stems from their fundamental architectural differences. FFLT is constrained to a single, preselected kernel, making its performance highly sensitive to that choice and vulnerable to suboptimal kernel selection. In contrast, OMKC optimally combines multiple kernel classifiers, which allows it to mitigate the risk of poor kernel choice and achieve more accurate predictions.

\item OMKC vs OKS-OMC:  
OMKC significantly outperforms OKS-OMC on F1, MacroF1 and MicroF1, with a near-significant edge on Psn. 
On Rcal, Hl and Rl, both algorithms are comparable. 
Overall, OMKC is superior to OKS-OMC. 
Its advantage can be attributed to two key innovations in its design. First, unlike OKS-OMC, OMKC employs a fixed label thresholding strategy for its single-kernel models. This design is advantageous in high-dimensional feature spaces induced by RFF techniques, where adaptively learning a label thresholding predictor carries the risk of overfitting.
Second, OMKC optimally combines multiple single-kernel models to leverage complementary information.
In contrast, OKS-OMC is designed to converge to the optimal single-kernel model. 
\end{itemize}

In summary, empirical results confirm that OMKC achieves superior overall performance compared to existing online multi-label classification methods.

\subsection{Sensitivity analysis}

In OMKC, the hyperparameters $\lambda$, $\eta$, and $\beta$ serve the same roles as in OKS-OMC. 
Since their sensitivity is theoretically identical to that in OKS-OMC, we omit a detailed analysis to avoid redundancy.
The number of RFF components, $D$, governs the trade-off between the approximation accuracy of kernel functions and the computational (space and time) cost of OMKC. A larger $D$ yields a better approximation, leading to improved performance, but at the expense of increased computation cost. We recommend selecting the largest $D$ possible while keeping the computational cost within an acceptable limit.

\begin{figure*}[htp]
	\centering
	\subfloat[]{\includegraphics[width=0.33 \textwidth]{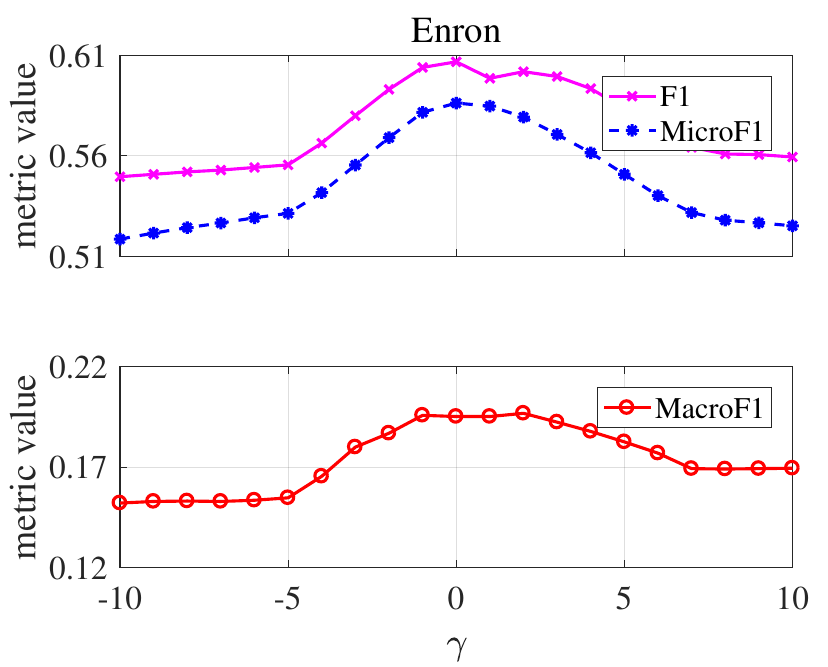}}
	\subfloat[]{\includegraphics[width=0.33 \textwidth]{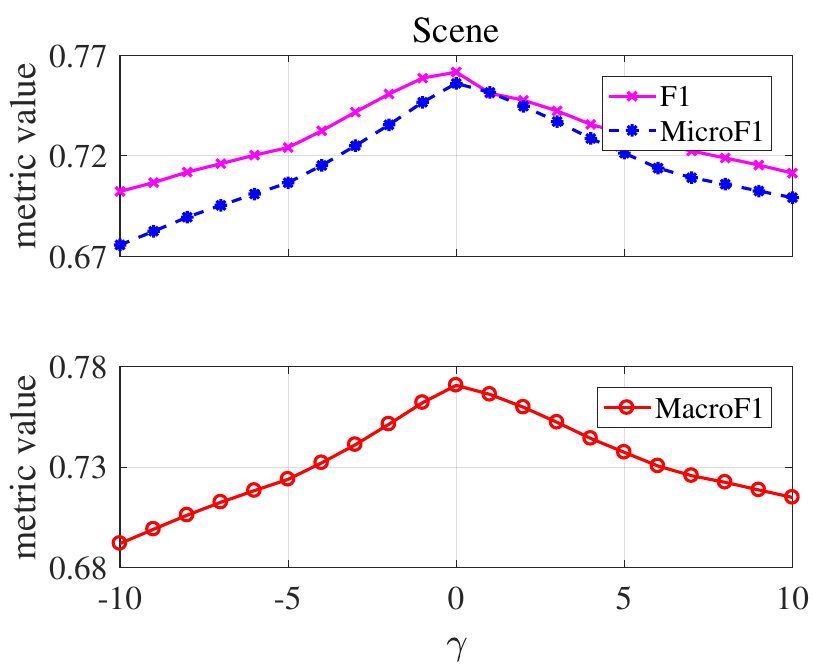}}
	\subfloat[]{\includegraphics[width=0.33 \textwidth]{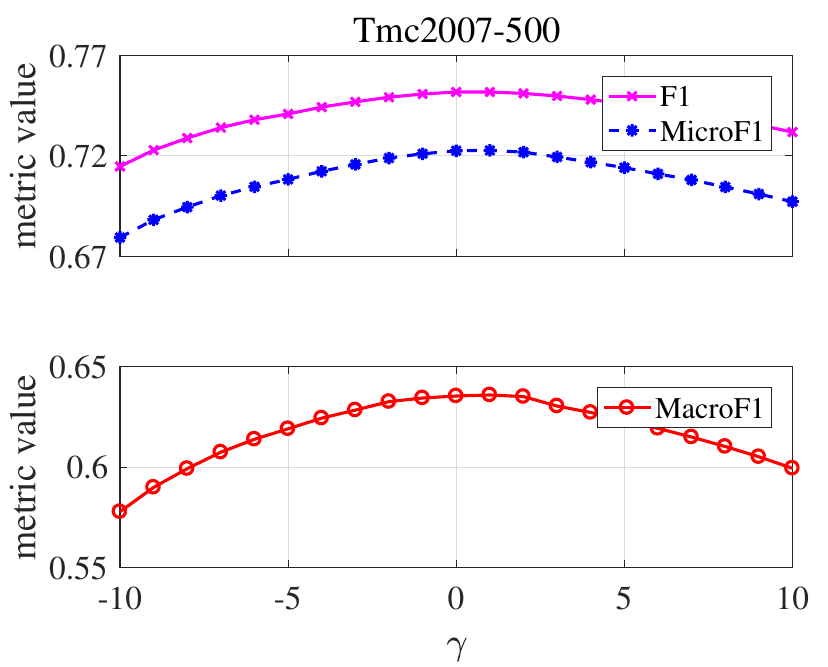}}

	\subfloat[]{\includegraphics[width=0.33 \textwidth]{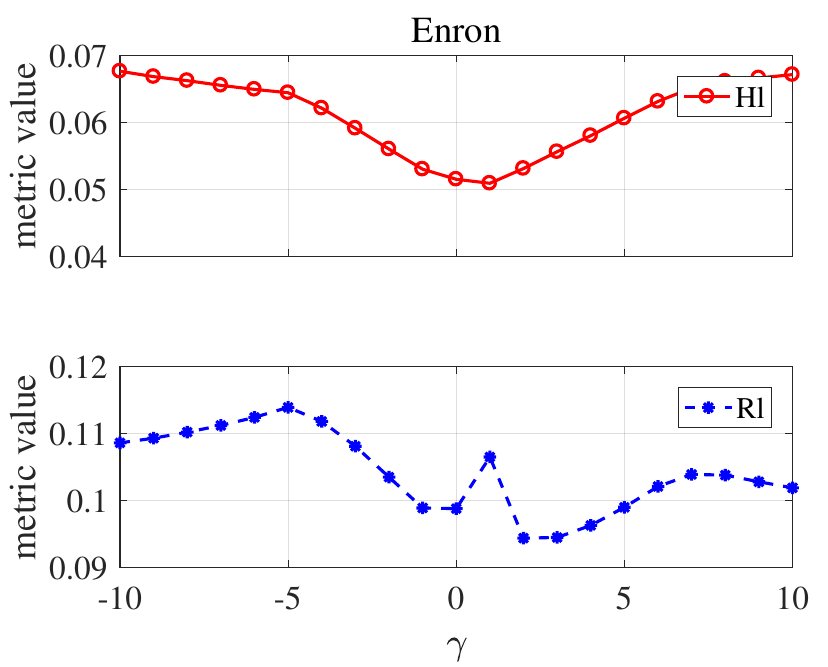}}
	\subfloat[]{\includegraphics[width=0.33 \textwidth]{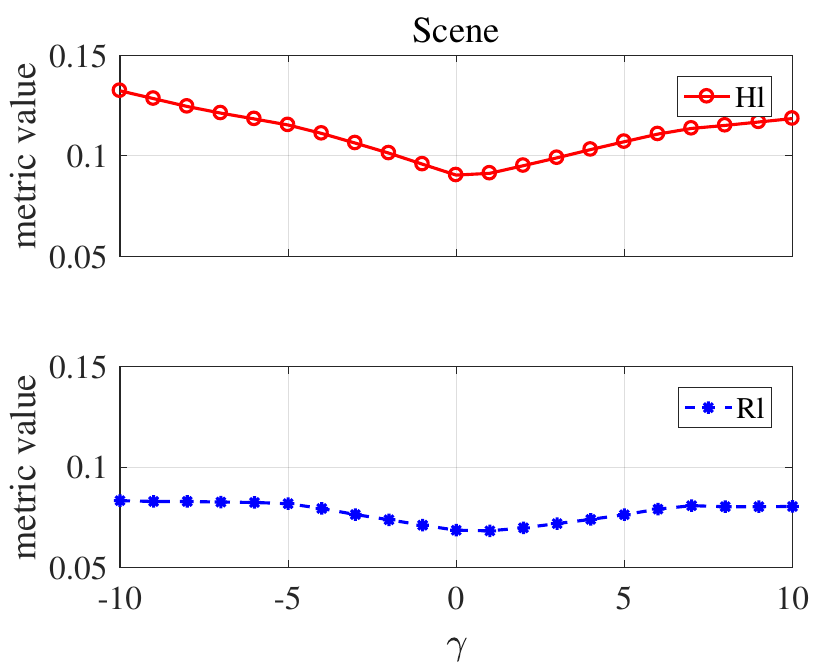}}
	\subfloat[]{\includegraphics[width=0.33 \textwidth]{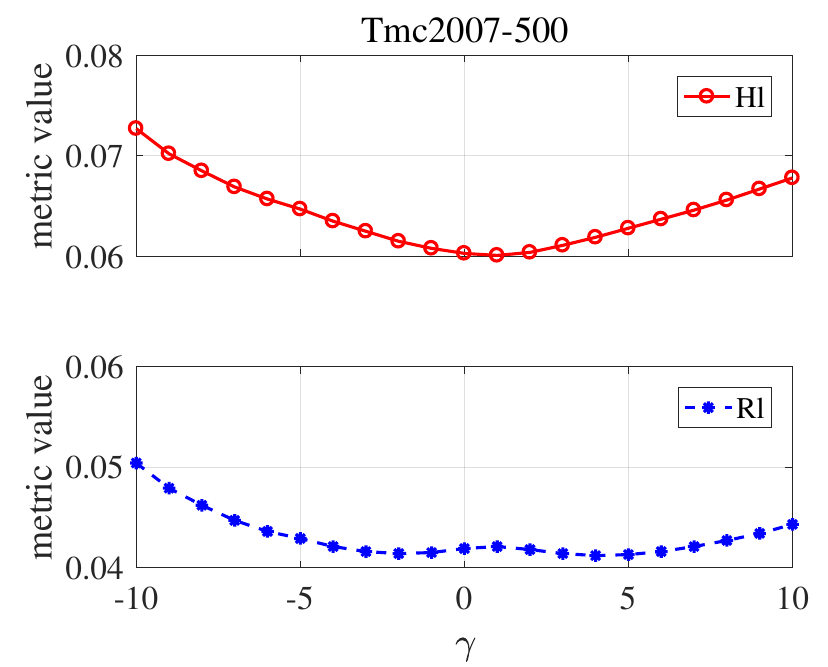}}
	\caption{Evolution of online performance metrics of OMKC as the fixed threshold $\gamma$ varies within the interval $[-10, 10]$. }
	\label{sensitivity}
\end{figure*}

The global fixed threshold $\gamma$ controls both the updating and prediction of the multi-kernel multi-label model $(\mathcal{W}, \bm \mu)$ in OMKC. We evaluate OMKC's sensitivity to this parameter using three randomly chosen datasets and five key metrics. Fig. \ref{sensitivity} illustrates how the online performance metrics of OMKC evolve as $\gamma$ varies within the interval $[-10, 10]$, while holding all other hyperparameters fixed.
From these figures, we observe that, across all three datasets and nearly all metrics, larger and smaller values of $\gamma$ lead to worse performance. The optimal $\gamma$ is consistently close to zero. Based on this observation, we recommend setting $\gamma = 0$ as the default value.

\section{Conclusions and limitations}
\label{sec.conclusion}

In this paper, we have introduced OMKC, an efficient online multi-kernel learning method for large-scale multi-label classification. By integrating advanced RFF techniques, OMKC effectively circumvents the curse of kernelization while maintaining high predictive performance. OMKC further improves robustness in high-dimensional RFF-induced feature spaces by employing a fixed label thresholding strategy in its single-kernel multi-label models.
In OMKC, the combination coefficients of classifiers are updated by directly minimizing the joint optimization objective function, while each kernel classifier is updated by minimizing an upper bound of this objective. 
This strategy enables rapid convergence to a high-quality solution in online learning scenarios.
We also provide a theoretical guarantee, proving that OMKC asymptotically converges to the optimal fixed convex combination of the best fixed single-kernel multi-label models in hindsight. 
Empirically, the superiority of OMKC was rigorously validated.
Ablation studies confirm the critical role of OMKC's coefficient update rule, especially when kernel approximation errors are large, and also show that—when coupled with this update rule—the fixed label thresholding strategy generally surpasses the adaptive one.
Moreover, evaluations across 11 datasets using seven metrics show OMKC statistically significantly outperforms state-of-the-art online multi-label classification methods in overall online performance.

Currently, this paper focuses primarily on stationary multi-label data streams; consequently, both the proposed OMKC and its theoretical analysis are confined to stationary settings. A promising direction for future research is to extend OMKC to handle non-stationary data streams \cite{FahyYG23}, where algorithms must adapt to concept drift over time. 
Another limitation lies in the sensitivity of OMKC to the choice of the predefined kernel pool. An ill-suited or overly restricted set of candidate kernels can significantly degrade performance, which underscores that selecting an appropriate kernel pool remains a critical consideration in practice.

\section*{Acknowledgements} 
This work was supported by the National Natural Science Foundation of China [grant number 61906165].


\setlength{\bibsep}{0pt}
\bibliographystyle{elsarticle_num}
\bibliography{reference.bib}

\end{document}